\documentclass[letterpaper, 10 pt, conference]{formatting/ieeeconf}
\IEEEoverridecommandlockouts                              
\let\labelindent\relax
\usepackage{amsfonts}       

\usepackage{amsthm}
\usepackage{mathtools}      
\usepackage{amssymb}        
\usepackage{filecontents}
\usepackage{graphicx}       
\usepackage{subcaption}
\usepackage{marginnote}     
\usepackage{marvosym}       
\usepackage{overpic}        
\usepackage{tabularx}
\usepackage{cite}
\usepackage{color}
\usepackage[linesnumbered,algoruled,boxed,lined]{algorithm2e}
\usepackage[normalem]{ulem}
\usepackage{epstopdf}
\usepackage{enumitem}
\usepackage[normalem]{ulem}
\usepackage{lipsum}
\usepackage{svg}
\usepackage[a-2b,mathxmp]{pdfx}[2018/12/22]
\usepackage{comment}

\newif\ifdraft
\draftfalse

\ifdraft
\usepackage[paperheight=11in,paperwidth=9.5in,
			left=1.25in,right=1.25in,
			top=0.75in,bottom=0.75in,
			heightrounded,marginparwidth=1.2in,
			marginparsep=0.05in]{geometry}
\usepackage{xcolor}
\usepackage{xargs} 
\usepackage[textsize=footnotesize]{todonotes}
\newcommandx{\jy}[2][1=]{\todo[linecolor=green,
			backgroundcolor=green!10,bordercolor=green,#1]{JY:#2}}
\newcommandx{\ax}[2][1=]{\todo[linecolor=blue,
			backgroundcolor=blue!10,bordercolor=blue,#1]{AX:#2}}
\newcommandx{\sw}[2][1=]{\todo[linecolor=orange,
			backgroundcolor=orange!10,bordercolor=orange,#1]{SF:#2}}
\newcommandx{\kg}[2][1=]{\todo[linecolor=red,
			backgroundcolor=red!10,bordercolor=red,#1]{Kai:#2}}
\else
\newcommand{\kg}[1]{{}}
\newcommand{\jy}[1]{{}}
\fi

\newif\iftwocolumn
\twocolumntrue

\newtheorem{problem}{Problem}
\newtheorem{proposition}{Proposition}[section]

\theoremstyle{definition}

\theoremstyle{remark}

\SetKwProg{Fn}{Function}{}{}
\SetKwComment{Comment}{$\triangleright$\ }{}

\makeatletter
\def\subsubsection{\@startsection{subsubsection}
                                 {3}
                                 {\z@ \hspace*{1mm}}
                                 {0ex plus 0.1ex minus 0.1ex}
                                 {0ex}
                                 {\normalfont\normalsize\itshape}}
\makeatother

\def\probm{\texttt{MORT}\xspace}
\def\algo{\texttt{SIPP}\xspace}
\def\b#1{\textcolor{blue}{#1}}

\title{\bf
Optimal and Stable Multi-Layer Object Rearrangement on a Tabletop
}
\author{
Andy Xu* \qquad Kai Gao* \qquad Si Wei Feng* \qquad Jingjin Yu
\thanks{
*These authors made equal contributions. 
All authors are with the Department of Computer Science, 
Rutgers, the State University of New Jersey, Piscataway, NJ, 
USA. 
This work is partly supported by NSF awards IIS-1845888 and IIS-2132972, and an Amazon Research Award.
}%
}

\begin{document}

\maketitle
\thispagestyle{empty}
\pagestyle{empty}


\begin{abstract}
Object rearrangement is a fundamental sub-task in accomplishing a great many physical tasks. As such, effectively executing rearrangement is an important skill for intelligent robots to master.  
In this study, we conduct the first algorithmic study on optimally solving the problem of \b{M}ulti-layer \b{O}bject \b{R}earrangement on a \b{T}abletop (\b{\probm}), in which one object may be relocated at a time, and an object can only be moved if other objects do not block its top surface. 
In addition, any intermediate structure during the reconfiguration process must be physically stable, i.e., it should stand without external support.
To tackle the dual challenges of untangling the dependencies between objects and ensuring structural stability, we develop an algorithm that interleaves the computation of the optimal rearrangement plan and structural stability checking. Using a carefully constructed integer linear programming (ILP) model, our algorithm, \b{S}tability-aware \b{I}nteger \b{P}rogramming-based \b{P}lanner (\b{\algo}), readily scales to optimally solve complex rearrangement problems of 3D structures with over $60$ building blocks, with solution quality significantly outperforming natural greedy best-first approaches. 

\vspace{2mm}
Upon the publication of the manuscript, source code and data will be available at \href{https://github.com/arc-l/mort/
}{\texttt{\textcolor{blue}{https://github.com/arc-l/mort/
}}}. \\
\end{abstract}

\section{Introduction}\label{sec:intro}
Object rearrangement is a sub-task of fundamental importance in accomplishing a great many physical tasks, be them in logistics, office spaces, or household settings. 
As such, performing fast, high-quality object rearrangement is an essential skill for the next generation of intelligent robots to master. 
%
Generally speaking, in a robotic rearrangement task, a manipulator must move a set of objects to reach some desirable, stable geometric configuration. 
%
Here, we examine optimally solving a subclass of rearrangement problems where objects to be organized are located on a \emph{tabletop}. 
In such setups, due to the unblocked access above the tabletop, it is reasonable to assume that relocating a single, unblocked object takes a constant amount of time, thus reducing the problem to a combinatorial \emph{task planning} challenge. 

Due to the intricate interactions among many objects, however, optimally solving the combinatorial task planning challenge in a tabletop rearrangement problem is highly non-trivial.
Indeed, it was formally proven that optimally rearranging a single layer of objects on a tabletop using pick-n-place manipulation primitives is NP-complete \cite{HanStiKonBekYu18IJRR} (in fact, the problem is APX-hard or hard to approximate.). 
\begin{figure}[h]
\vspace{2mm}
     \centering
    \includegraphics[width=\columnwidth]{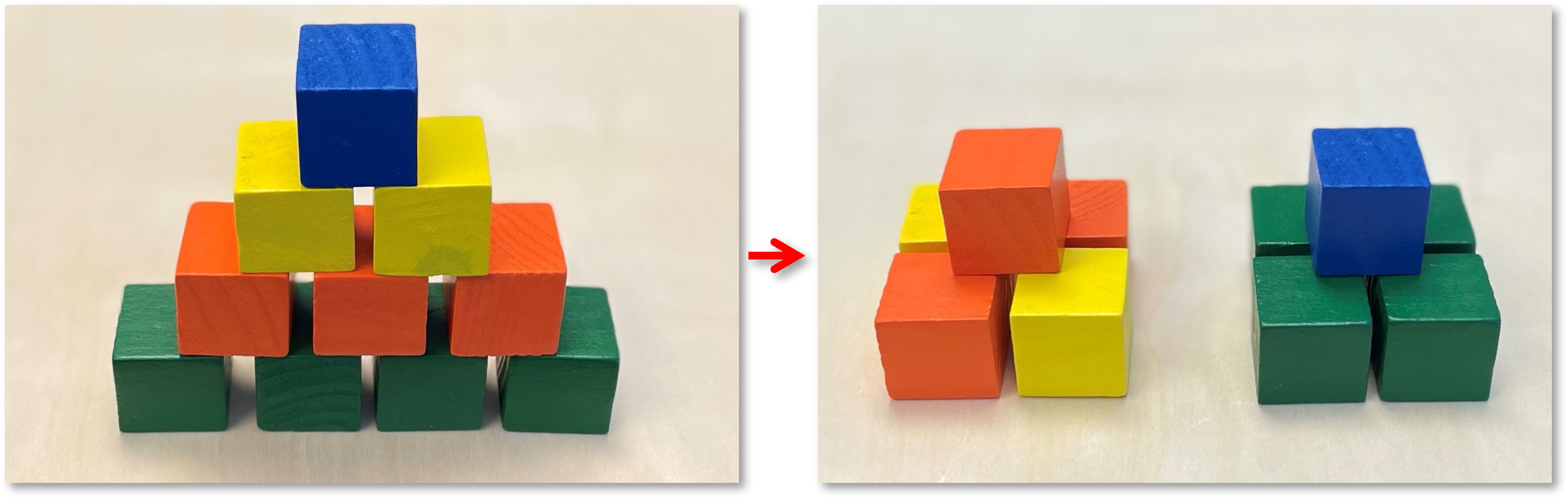}
    \caption{Illustration of a multi-layer rearrangement problem studied in this paper, where $10$ (uniquely labeled) blocks must be reconfigured from one stable arrangement to another stable arrangement through a sequence of stable intermediate arrangements, using a single robotic manipulator that can relocate one object at a time.}
    \label{fig:ex}
\end{figure}
Roughly speaking, the difficulty comes from the overlap of start and goal configurations of objects, which leads to intricate \emph{dependencies} among the objects. 
Minimizing the number of rearrangement actions to solve a tabletop rearrangement problem amounts to untangling the dependencies. Despite the computational intractability, multiple methods \cite{HanStiKonBekYu18IJRR,GaoFenYu21RSS,GaoLauHuaBekYu22ICRA,gao2023minimizing} have been developed to quickly compute high-quality solutions according to multiple metrics.
%
%

With the structure of the single-layer tabletop rearrangement problem reasonably well understood, we ask a natural next question: what if the start and goal configurations contain multiple layers of objects? 
In this work, we study the Multi-layer Object Rearrangement on a Tabletop (\probm) problem, in which the start/goal configurations may contain many objects stably organized in many layers. 
An object may be relocated if other objects do not occupy its top surface. A relocation must also be stable, i.e., an object, after being released at a target pose, is stably supported by the tabletop or other objects. 
As an abstract model,
in addition to modeling multi-layer rearrangement at the tabletop scale, \probm can be applied to large-scale problems, e.g., it can be readily combined with motion planning methods in \cite{9868234} to accelerate the construction of large structures.


Similar to the solutions structure of single-layer tabletop rearrangement problems, which is heavily coupled to the dependency graph of the objects, object dependencies are also key to solving \probm. 
Because the single-layer setting is a special case and is NP-hard, \probm is NP-hard as well. 
However, \probm must also deal with layer-based object dependencies.  
Moreover, solving \probm requires considering not only the combinatorial dependencies among the objects but also structural stability, which limits the number of possible construction pathways. %

To tackle combinatorial constraints and structural stability jointly, we propose \algo (Stability-aware Integer Programming-based Planner), an optimal algorithm that interleaves solving an integer linear programming (ILP) model and checking structural stability via physics-based simulation.
Specifically, we first solve an ILP model for the shortest rearrangement plan fulfilling all combinatorial constraints. Then, the stability of all intermediate arrangements is checked in a physics simulator. If some arrangements are unstable, the failure is added as linear constraints to the ILP model to refine the plan. The process iterates until a feasible plan is found, which is optimal.
Thorough simulation evaluations demonstrate that \algo has excellent scalability and, as an optimal algorithm, outperforms best-first approaches by a large margin in terms of the number of objects that must be temporarily relocated, which translates to big potential savings in execution time. 

In summary, our study brings two main contributions. First, we initiate the structural and algorithmic study of the Multi-layer Object Rearrangement on a Tabletop (\probm) problem, a task planning problem modeling the optimal transformation of 3D structures with applications at multiple domains and scales, e.g., making the large-scale robotic constructions more efficient.  
Second, we develop a novel and optimal algorithm for \probm, by interleaving solving an optimal integer linear programming model and introducing stability-assurance linear constraints. The iterative solution framework, taking advantage of the structure of the \probm problem, readily scales to perform optimal in-place rearrangement of complex 3D structures with over $60$ building blocks, with solution quality significantly better than (greedy) best-first methods.


\textbf{Paper Organization}
The remainder of this paper is structured as follows. Sec.~\ref{sec:related}, introduces previous works on related topics and contrasts them with this work. 
Sec.~\ref{sec:problem}, presents the formal definition of \probm and some of its structural properties.
Sec.~\ref{sec:algorithm} describes \algo, an optimal algorithm for \probm, along with a greedy best-first search baseline.
Section~\ref{sec:evaluation} evaluates the performance of \algo, and finally, Sec.~\ref{sec:conclusion} discusses and concludes the paper.

\section{Related Work}\label{sec:related}
\textbf{Object Rearrangement}.
Robotic rearrangement, as a subcategory of task and motion planning problems with a vast number of applications, has been extensively studied. Many challenging setups have been examined, including dense single-layer tabletop settings \cite{GaoYu22IROS,danielczuk2021object,ShoSolYuHalBek18WAFR}, shelves that constrain robots' motion \cite{WanGaoYuBek22ICAPS,wang2022efficient,wada2022reorientbot}, and furniture rearrangement in a large workspace \cite{gu2022multi,gan2022threedworld,szot2021habitat}.
In tabletop rearrangement, robots can use both simple non-prehensile actions like pushing and poking \cite{huang2019large,Han21CASE,song2020multi,HuaHanBouYu21ICRA},
as well as more complex prehensile actions like grasping to manipulate objects and facilitate long-horizon planning \cite{wang2022efficient, GaoYu22IROS,gao2022utility}.
The combinatorial aspect of tabletop rearrangement poses significant challenges.
Previous works have used dependency graphs to encode combinatorial constraints and represent object relationships\cite{HanStiKonBekYu18IJRR,GaoFenYu21RSS,GaoLauHuaBekYu22ICRA,gao2023minimizing}. 
Most prior research in this area focuses on single-layer settings, where no object is blocked from above. 
Through dependency graphs, a single-layer tabletop rearrangement problem can be transformed into well-studied graph problems.
\probm, studied in this paper, extends the combinatorial study of rearrangement to multi-layer setups. 
Like \probm, stack rearrangement \cite{HanStiBekYu18RAL,SzeYu21SPAR}  also examines multi-layer object rearrangement from a combinatorial perspective, but with the stricter restriction that objects are stored in stacks, forming linear top-down dependencies. 

%

\textbf{Blocks World and Symbolic Reasoning}. Blocks World\cite{russell2010artificial} is a classical problem that received extensive attention in symbolic reasoning research. The development of this line of work initially focused mainly on logical reasoning, leading to the appearance of tools including  STRIPS (Stanford Research Institute Problem Solver)\cite{fikes1971strips} and PDDL (Planning Domain Definition Language)\cite{aeronautiques1998pddl}. Recently, a seminal work \cite{garrett2020pddlstream} 
integrates PDDL with motion planning to deliver PDDLStream, a general task and motion planner.
In contrast to these studies, our study of \probm focuses on taming the combinatorial explosion of a more complex version of the Blocks World problem, with the goal of solving it with the least number of operations. 

\textbf{Assembly and Disassembly Planning}
Assembly and disassembly planning are well-established fields in object manipulation with applications across a broad spectrum of domains. 
Earlier works in these areas focused on constructing symbolic reasoning systems to find assembly/disassembly sequences that satisfy given constraints\cite{de1987simplified,de1989correct,de1990and}. 
For example, Mello et al.\cite{de1989correct} used a relational model graph to represent part relationships and identify assembly sequences. 
To enable autonomous assembly, some studies have investigated geometric reasoning for assembly tasks\cite{wilson1994geometric,jimenez2013survey}, directly reasoning assembly sequences from component geometries. 
Previous research on assembly planning has also considered constraints like manipulation feasibility ('graspability')\cite{wan2018assembly,dobashi2014robust}, structural stability\cite{wan2018assembly,mcevoy2014assembly}, and coordination among manipulators\cite{dogar2019multi}. 
To ensure stability during assembly, Wan et al.\cite{wan2018assembly} evaluated 'stability quality' based on the object's center of mass and its supporting region. 
In related problems, Noseworthy et al.\cite{noseworthy2021active} proposed a GNN model for the stability of stacks of cuboid objects, while Garrett et al.\cite{garrett2020scalable} used the fixed-end beam equation to approximate the potential deformation of a spatial frame structure. 
Most assembly/disassembly problems assume that parts are scattered in the workspace in the start or the goal configuration. 
In contrast, \probm involves compact piles of objects in both the start and goal configurations. 
Additionally, previous stability checkers assume specific object shapes; we use a physics engine to evaluate stability for objects with general bases.

\section{Preliminaries}\label{sec:problem}
\subsection{Problem Formulation}
\begin{figure}[h]
    \centering
        \includegraphics[height=0.9in]{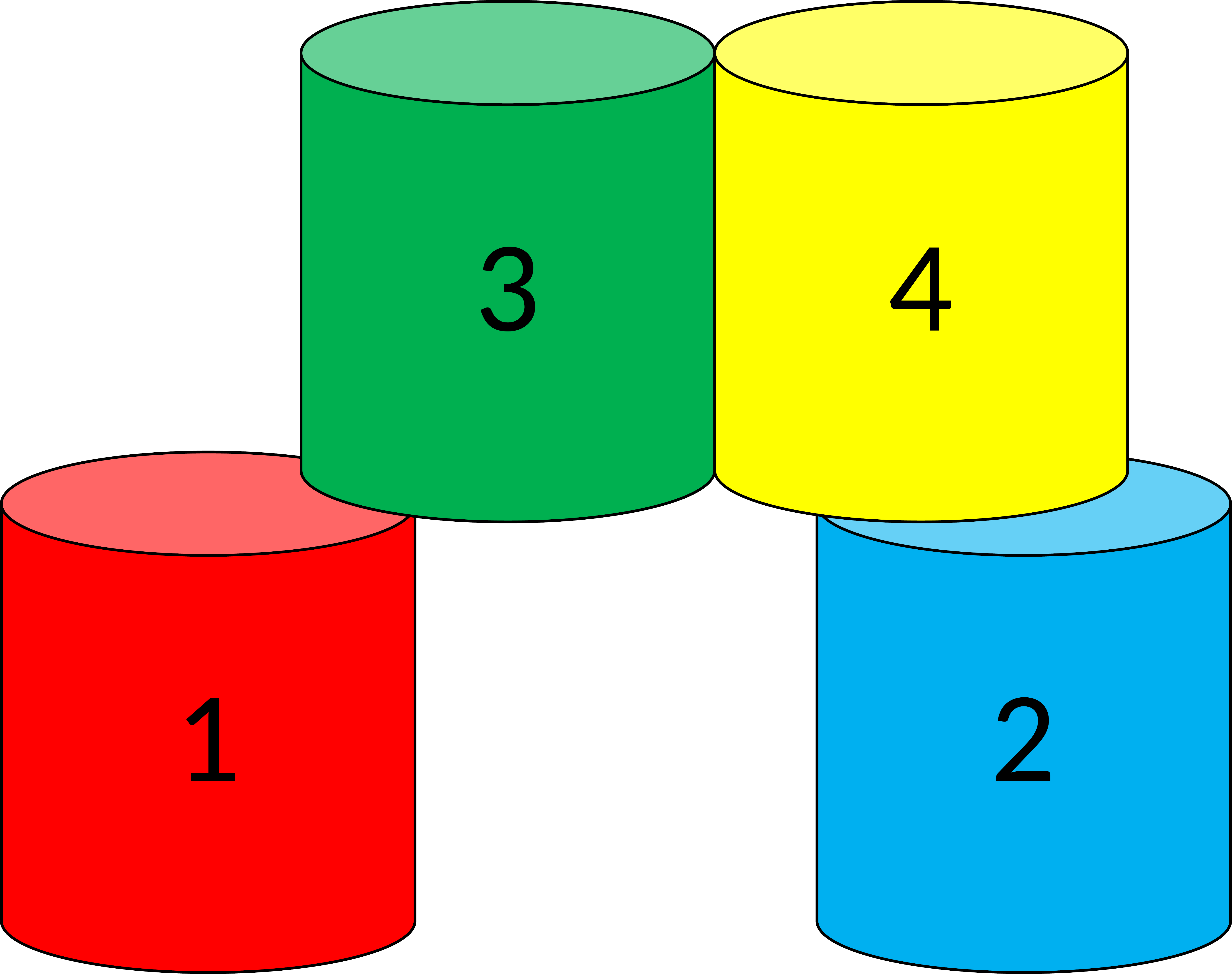}
    \caption{An arrangement of four objects. The arrangement is stable (assuming sufficient friction). If object $4$ is not present, the leftover arrangement is unstable; object $3$ will fall. 
    \label{fig:stability}}
\end{figure}

Consider a tabletop workspace with sufficient clearance of $H$ above it. Denote the tabletop as $\mathcal W_t \subset \mathbb R^2$; the workspace is then $\mathcal W \subset \mathcal W_t \times [0, H]$.
A set of $n$ objects $O = \{o_1, \dots, o_n\}$ is placed in $\mathcal W$, supported on the tabletop $\mathcal W_t$.
The pose $x_i \in \mathcal SE(3)$ of an object $i, 1\le n$ is \emph{stable} if the object remains motionless without external force.
As an example, the arrangement of four objects shown in Fig.~\ref{fig:stability} is stable. However, if object $4$ is not present, then the leftover arrangement is \emph{unstable} because object $3$ is not well-supported by object $1$.

%
A feasible arrangement of $O$ is a set of poses $A = \{x_1, \dots, x_n\}$, where each $x_i \in SE(3)$  is collision-free and stable.
In a single \emph{pick-n-place} rearrangement operation, an object at a pose without any overhead blockage may be relocated to another such pose. For example, this can be achieved using a suction cup-based gripper.   
%
A pick-n-place operation is formally represented as a triple $a = (i, x_i', x_i'')$, denoting that object $i$ is moved from pose $x_i'$ to pose $x_i''$.
We seek a plan $P = (a_1, a_2, \dots)$, an ordered sequence of pick-n-place operations, that moves objects from a start arrangement $A_1$ to a goal arrangement $A_2$. 
We say that $P$ is \emph{feasible} if all intermediate arrangements based on $P$ including $A_1$ and $A_2$ are feasible. 
Intermediate arrangement stability is necessary for a single robotic manipulator to execute the plan.



It is clear that in a feasible plan $P$ that moves objects from $A_1$ to $A_2$, for some object $i$ with $x_i' \in A_1$ and $x_i'' \in A_2$, $P$ may have to move $i$ from $x_i'$ to some temporary pose before moving it to $x_i''$. 
We may view a pick-n-place that moves an object to a temporary pose as moving the object to a \emph{buffer}. A buffer may be internal or external to $\mathcal W$; in  this work, a sufficient number of buffers is assumed to be available on $\mathcal W_t$. Nevertheless, it is generally desirable to use as few buffers as possible. 
With the introduction of buffers, three types of pick-n-place operations may happen to an object $i$ with $x_i' \in A_1$ and $x_i'' \in A_2$: 1) relocate $i$ directly from $x_i'$ to $x_i''$, 2) relocate $i$ from $x_i'$ to buffer pose, and 3) relocate $i$ from a buffer pose to $x_i''$.

It is generally preferred to solve a rearrangement problem with the least number of pick-n-places, which leads to the following natural rearrangement planning problem.

\begin{problem}[Multi-layer Object Rearrangement on a Tabletop (\probm)]
Given feasible start and goal arrangements $A_1$ and $A_2$, compute a feasible rearrangement plan $P$ as a sequence of pick-n-places with minimum $|P|$.
\end{problem}

As an initial study of \probm, in this work, we assume that all objects are \emph{generalized cylinders} with the same height. In other words, each object may be viewed as extending a connected and compact two-dimensional shape by some fixed height $h$. The assumption mirrors practical scenarios, e.g., constructing structures with bricks. Examples from \cite{9868234} also mainly address such types of objects.  

\subsection{Basic Structural Properties of \probm}
As mentioned, \probm, as a three-dimensional rearrangement challenge, induces two types of dependency-based combinatorial constraints. First, an object cannot be moved from its start pose if it is blocked from the above, or its removal leaves an unstable arrangement. Similarly, an object cannot be moved to the goal if its goal pose is blocked or not well-supported. For example, in the arrangement given in Fig.~\ref{fig:stability}, object $1$ cannot be moved without moving object $3$ first as object $3$ is above it. Object $4$ cannot be moved without moving object $3$ first because removing object $4$ first leaves an unstable arrangement. We call this first type of constraint \emph{contact} constraints. Contact constraints can add to the number of required pick-n-place operations for solving a \probm instance. 

The dependencies between start and goal arrangements induce the second type of constraint. For example, without considering stability, if we are to exchange the poses of Objects $3$ and $4$ in the arrangement given in Fig.~\ref{fig:stability}, one of them must be moved to a buffer (pose) before the task can be solved. As a result, to exchange the pose of two objects, at least three pick-n-place operations are needed, including one that moves an object to a buffer. We call this second type of constraints \emph{cyclic} constraints since they are induced by \emph{mutual} dependencies between objects.

If there is a single layer of objects, then \probm degenerates to the Tabletop Object Rearrangement with Overhand grasps (TORO) problem, where cyclic constraints make the
problem NP-complete~\cite{HanStiKonBekYu18IJRR}. As a result, i.e., TORO is a special case of \probm, \probm is also NP-complete. 

\begin{proposition}
\probm is NP-complete. 
\end{proposition}

In optimally solving \probm, both contact and cyclic constraints must be properly addressed. The NP-hardness of \probm suggests that no polynomial-time algorithms exist for exactly solving \probm unless P$=$NP. 



\section{Algorithms}\label{sec:algorithm}
Given the NP-completeness of \probm, we explored multiple natural approaches including dynamic programming (DP) and integer linear programming (ILP). Out of what we attempted, ILP stands out as the most promising, capable of computing optimal solutions fairly quickly. As a result, we report here a unique ILP solution we developed for solving \probm. We also describe a fast greedy best-first approach, which we implemented as a natural comparison point.

We call the ILP based algorithm 
Stability-aware Integer Programming-based Planner (\algo).
\algo first considers the case when all possible intermediate stages are stable and introduces an ILP model describing it.
Then, a subroutine checks the stability of the solution returned by the ILP model.
The above two subroutines iterate to give the final optimal solution.
In what follows, we describe details of \algo, followed by a brief description of the greedy method. 

\subsection{Stability-aware Integer Programming-based Planner}
\subsubsection{Permutation Based Integer Programming}
We represent the order of object removal from their start pose using a permutation 
\[
\tau = \begin{pmatrix}
1 & \dots & n\\
t_1 & \dots & t_n
\end{pmatrix}
\]
where object $i$ is the $t_i$-th object being moved out of its start pose. 
These $t_i$'s will appear as integer variables in the ILP model.
We may enforce that $t_1, \dots, t_n$ is a permutation of $\{1, \dots, n\}$, by ensuring $1\leq t_i \leq n$ for all $i$, and $t_i\neq t_j$ for all $i \neq j$ (this is necessary in the ILP model).
In addition, we have binary variables $b_1,\dots, b_n$,
where $b_i$ indicates whether object $i$ needs to be moved to a buffer.

We construct three sets of constraints to characterize the sequence.
First, if object $i$ is directly above object $j$ in the start configuration, then
\begin{equation}\label{eq:1}
    t_i < t_j. 
\end{equation}

Second, for all pairs of $(i, j)$ that object $i$ is above object $j$ in the goal configuration
if $t_i < t_j$, which means object $i$ is moved out of the start configuration before object $j$,
object $i$ must be put into a buffer as it cannot be put into the goal as there exists object $j$ that is supposed to be below it, which has not been retrieved from its start pose yet.
\begin{equation}
\label{eq:2}
    (t_i < t_j) \rightarrow b_j.
\end{equation}

Lastly, for all pairs of $(i,j)$ that object $i$'s goal is in collision with object $j$'s start, 
then if $p_i < p_j$, object $i$ must be put into a buffer.
\begin{equation}
\label{eq:3}
   (t_i < t_j) \rightarrow b_j.
\end{equation}

Naturally, as the total number of moves is $n+\sum_i b_i$, we have the summation of $b_i$ as the objective 

\begin{equation}
    \min \sum_{i=1}^{n} b_i
\end{equation}

After solving the ILP model, a valid permutation $\tau$ is obtained, from which we can derive the corresponding plan by computing $\tau^{-1}$ as 

\begin{equation}\label{eq:5}
\begin{pmatrix}
1 & \dots & n\\
p_1 & \dots & p_n
\end{pmatrix}    
 = \tau^{-1} = \begin{pmatrix}
t_1 & \dots & t_n\\
1 & \dots & n
\end{pmatrix}.
\end{equation}

And $b_j$ indicates the buffer usage of object $j$. 
It is straightforward to see that the three types of constraints are necessary. We now show they are also sufficient. 
Suppose that we follow the sequence computed by $p_1, \dots, p_n$ to move the objects out of the buffer.

First of all, all the objects are able to be moved out of the start pose since all its upper objects are moved out 
because of the constraint given in Eq.~\eqref{eq:1}.
Then, we only need to show if $b_i=0$, object $i$ can be moved to the goal directly, because if $b_i=1$, object $i$ will be moved to a buffer which can be executed without issue.
By constraint Eq.~\eqref{eq:2}, if $b_i=0$, all objects below it in the goal pose have already been moved out of their start poses. 
And by constraint Eq.~\eqref{eq:3}, if $b_i=0$, all objects whose initial poses are in collision with $i$'s goal have already been removed from its start. 
Hence, all objects that are supposed to be under object $i$ in the goal configuration can be already put into the goal before we manipulate object $i$, and this means object $i$ can be put into the goal directly.
\vspace{1mm}

\subsubsection{Checking Stability of Intermediate States}
\begin{figure}[!ht]
    \centering
    \begin{subfigure}{.22\textwidth}
    \centering
        \includegraphics[height=0.9in]{figures/unstable-case-start_svg-raw.pdf}
        \caption{Start pose}
        \label{}
    \end{subfigure}
    \hspace{.1in}
    \begin{subfigure}{.22\textwidth}
    \centering
        \includegraphics[height=0.9in]{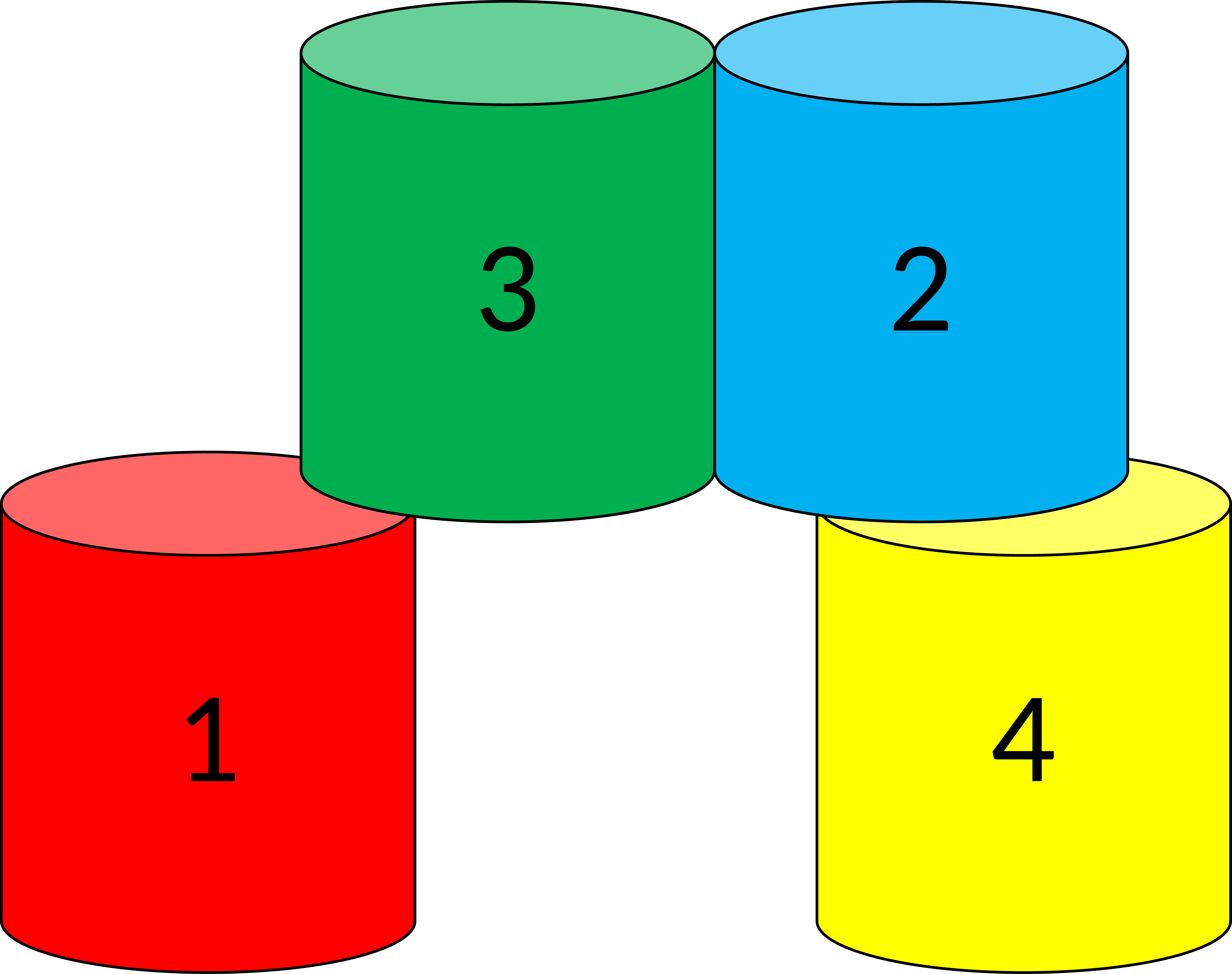}
        \caption{Goal pose}
        \label{}
    \end{subfigure}
    \caption{A \probm instance where a plan does not consider arrangement stability will fail to yield a feasible plan. An optimal plan requires six pick-n-places, moving each of objects $2$-$4$ twice.\label{fig:1}}
\end{figure}

It is possible that the ILP model yields solutions that are unstable.
For example, when rearranging from Fig.~\ref{fig:1}(a) to Fig.~\ref{fig:1}(b), the plan generated by solving the ILP model moves objects 4 and 2 to buffers, and then to their goal poses.
However, moving object 3 destabilizes object 2, but a stable solution exists where 2 is moved to a buffer first.

\begin{figure}[!ht]
    \centering
    \begin{subfigure}{.22\textwidth}
    \centering
        \includegraphics[height=0.9in]{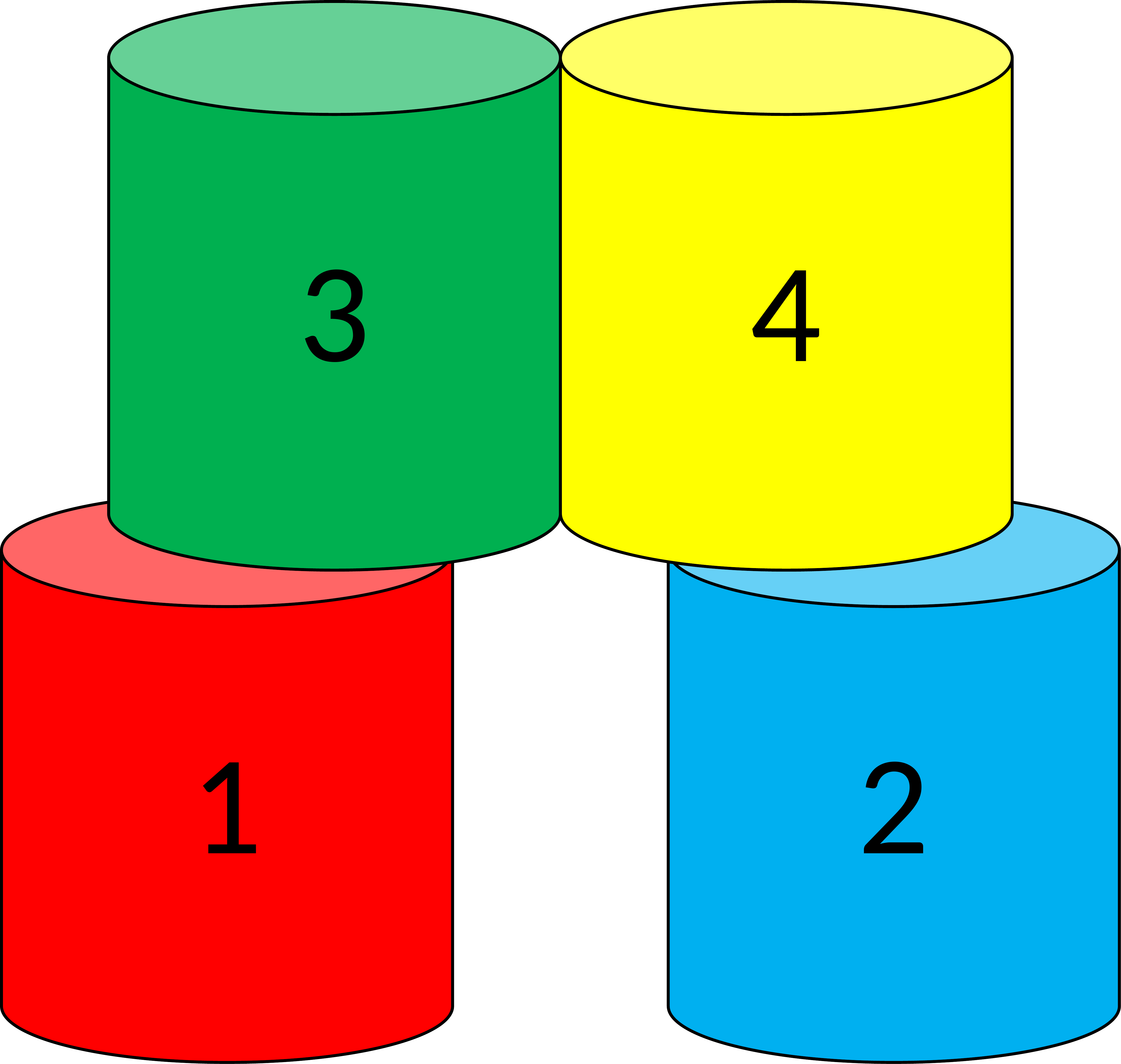}
        \caption{Start pose}
        \label{}
    \end{subfigure}
    \hspace{.1in}
    \begin{subfigure}{.22\textwidth}
    \centering
        \includegraphics[height=0.9in]{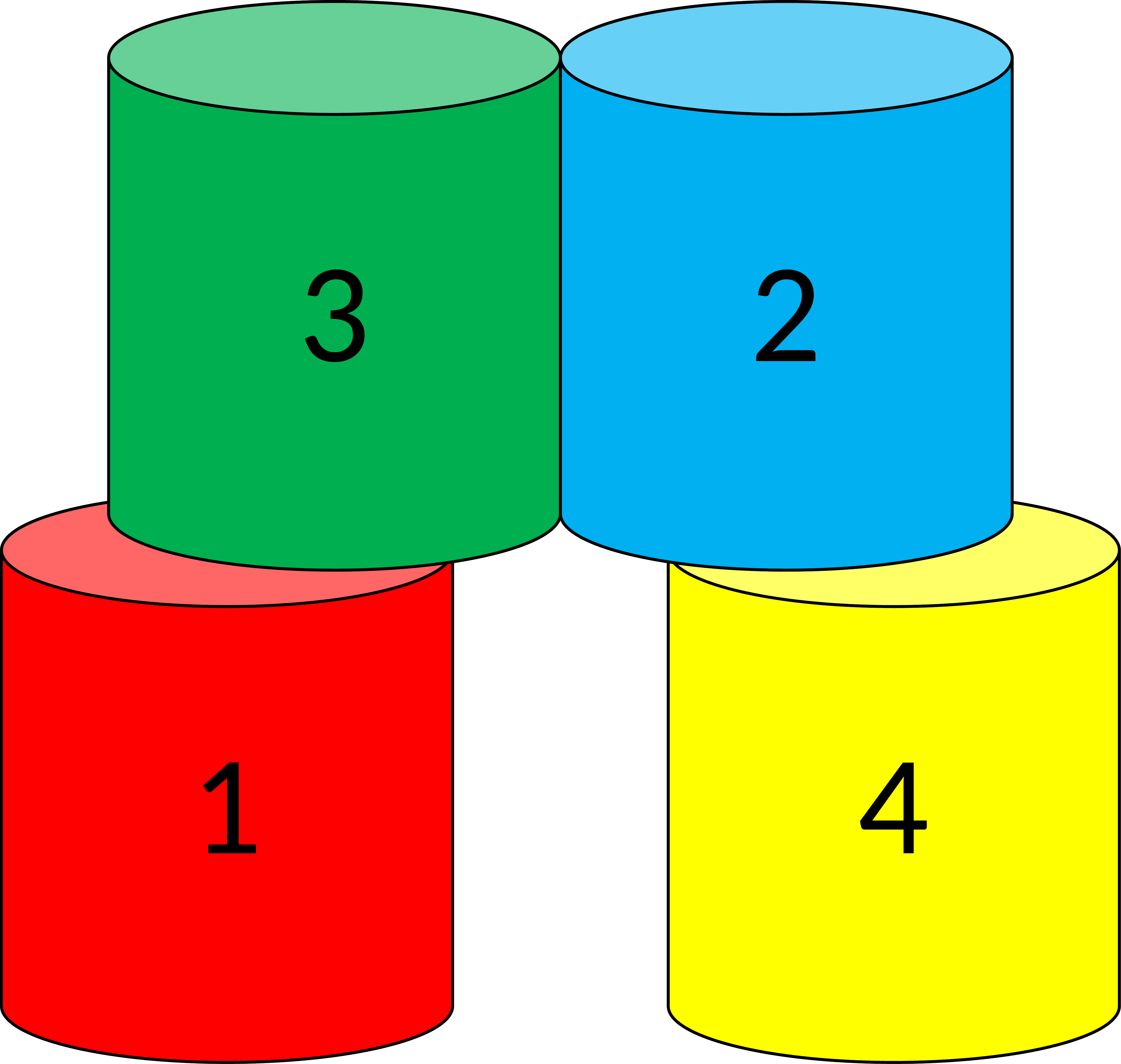}
        \caption{Goal pose}
        \label{}
    \end{subfigure}
    \caption{A \probm instance that is close to that in Fig.~\ref{fig:1} but can be solved using $4$ pick-n-places. Intermediate arrangements with object $2$ or object $3$ missing from the goal configuration are both stable. \label{fig:2}}
\end{figure}

\begin{figure}[!ht]
\vspace{2mm}
    \centering
    \begin{subfigure}{.22\textwidth}
    \centering
        \includegraphics[height=0.9in]{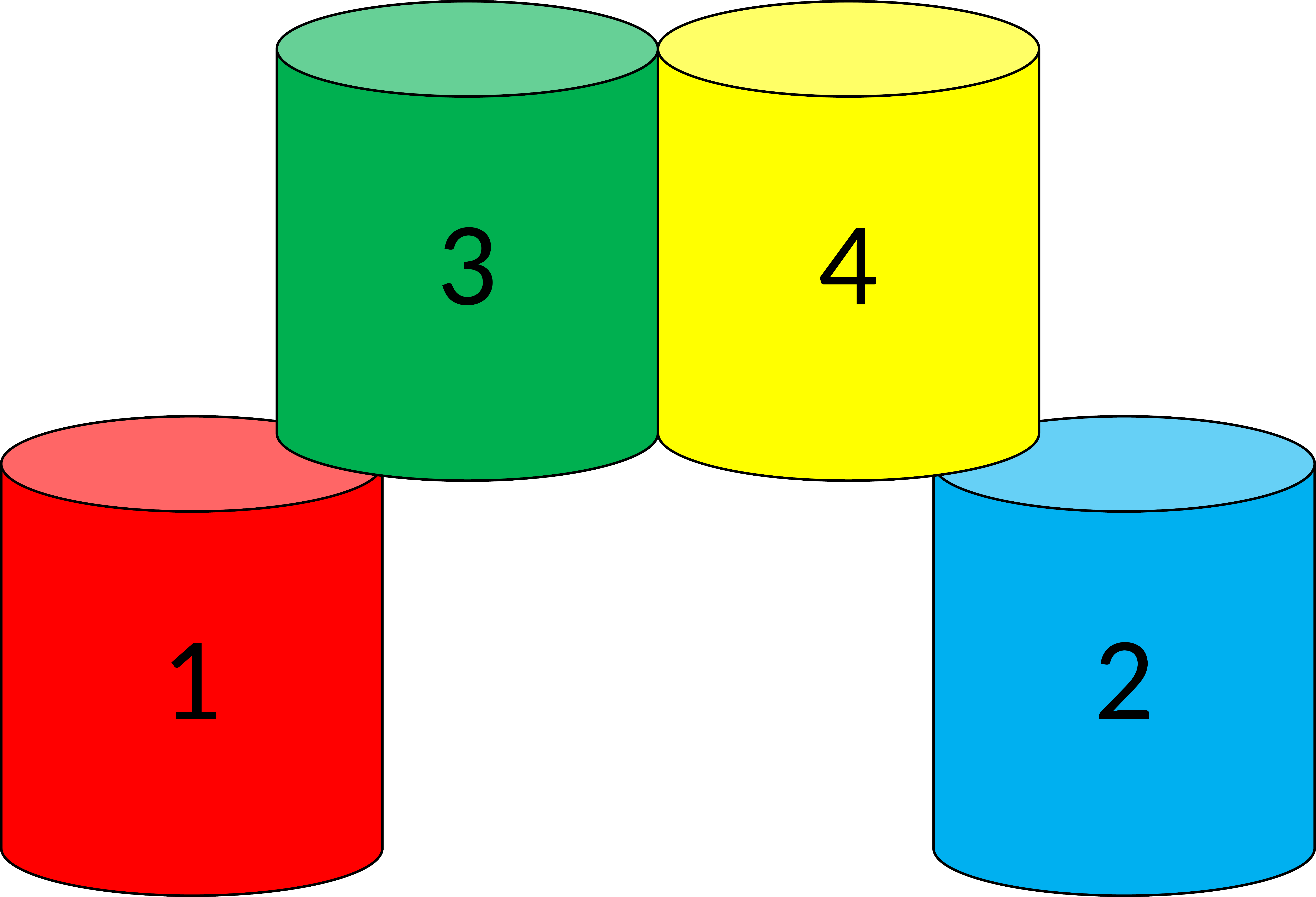}
        \caption{Start pose}
        \label{}
    \end{subfigure}
    \hspace{.1in}
    \begin{subfigure}{.22\textwidth}
    \centering
        \includegraphics[height=0.9in]{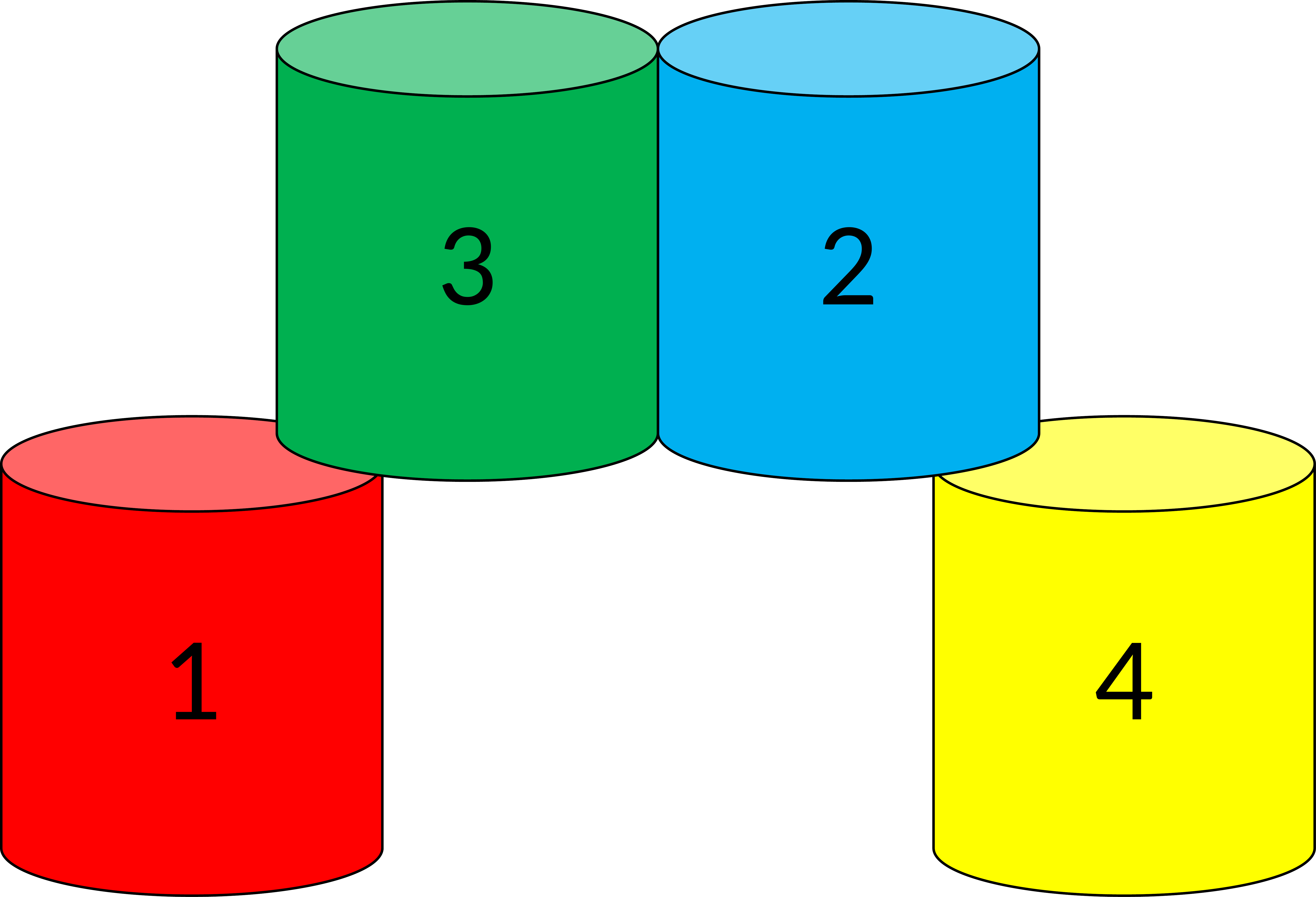}
        \caption{Goal pose}
        \label{}
    \end{subfigure}
    \caption{An infeasible \probm instance involving $4$ objects, because there does not exist a stable intermediate arrangement with three objects at their goal poses. \label{fig:3}}
\end{figure}

Furthermore, there exists cases with the same structural representations as Fig.~\ref{fig:1}, but have no such stability limitations. One such case is shown in Fig.~\ref{fig:2}.
Similarly, there exists cases with the same structural representations as Fig.~\ref{fig:1} and Fig.~\ref{fig:2}, but are infeasible under the rules of the problem, such as the one shown in Fig.~\ref{fig:3}. This is because moving any of the 2 objects in the top layer will destabilize the structure as the two objects are supported due to the mutual friction between them.

A simple way to check arrangement stability is through a physics simulator. We use pyBullet \cite{coumans2019} to simulate the actions of the robot along each step of the solution produced by the ILP model. After each move, we run the simulation for a fixed amount of time to check if any objects are unstable. Stability is determined by checking if the distance between each object's actual and expected locations exceeds a certain threshold. If any objects are found to be unstable, the violating move is returned.

\vspace{1mm}

\subsubsection{Integration}
The integration of the ILP modeling and stability checking is straightforward. In case a sequence $\{\bar{t}_i\}_{i=1}^{n}$ is not stable at the $(m+1)$-th operation, the following constraint is added
\begin{equation}
 (\{\bar p_1, \dots, \bar p_m\} = \{p_1, \dots, p_m\}) \rightarrow (p_{m+1} \neq \bar{p}_{m+1}),
\end{equation}
which is equivalent to
\begin{equation}
 [(t_{\bar{p}_1} \leq m) \wedge \dots \wedge (t_{\bar{p}_m} \leq m)] \rightarrow (t_{\bar{p}_{m+1}} \neq \bar{t}_{\bar{p}_{m+1}}).
\end{equation}

In other words, if the variables $p_1,...,p_m$ all move before the $m$th operation, they will reach the same state as before at the $m$th operation, so the next move must not be the same as the move previously found to be unstable.

\subsection{Greedy Best-First Algorithm}
We implement the greedy best-first algorithm for \probm as follows. 
First, start pose dependencies are processed in topological order, from the bottom layer to the top layer. 
%
%
For each object $i$ that appears in the list of objects in the topological ordering, $i$'s goal pose is freed by moving all objects in the start pose for which $i$ has a contact constraint (or equivalently, dependency), in a recursive manner (i.e., if an object $j$ blocks $i$ from the above, then all objects blocking $j$ must also be removed). 
If these objects can be moved to goals directly, we do so.  
Otherwise, they are moved to buffers. 
Algorithmically, this is done using breadth-first search (BFS) on the start pose constraint structure from all objects $o_{s_j}$ in the set of objects $s$ such that $o_{s_j}$'s start pose directly collides with $i$'s goal pose.
If $i$ is in its start pose, free it by moving all objects above $i$ in the start pose to their goals if they're available. Otherwise, these objects are relocated to buffers.
Then, $i$ is moved directly to its goal.
If any objects in buffers can be moved to their respective goal poses at any point, we do so because this is the best that one can do.

\begin{figure}[!ht]
    \centering
    \begin{subfigure}{.22\textwidth}
    \centering
        \includegraphics[height=1.3in]{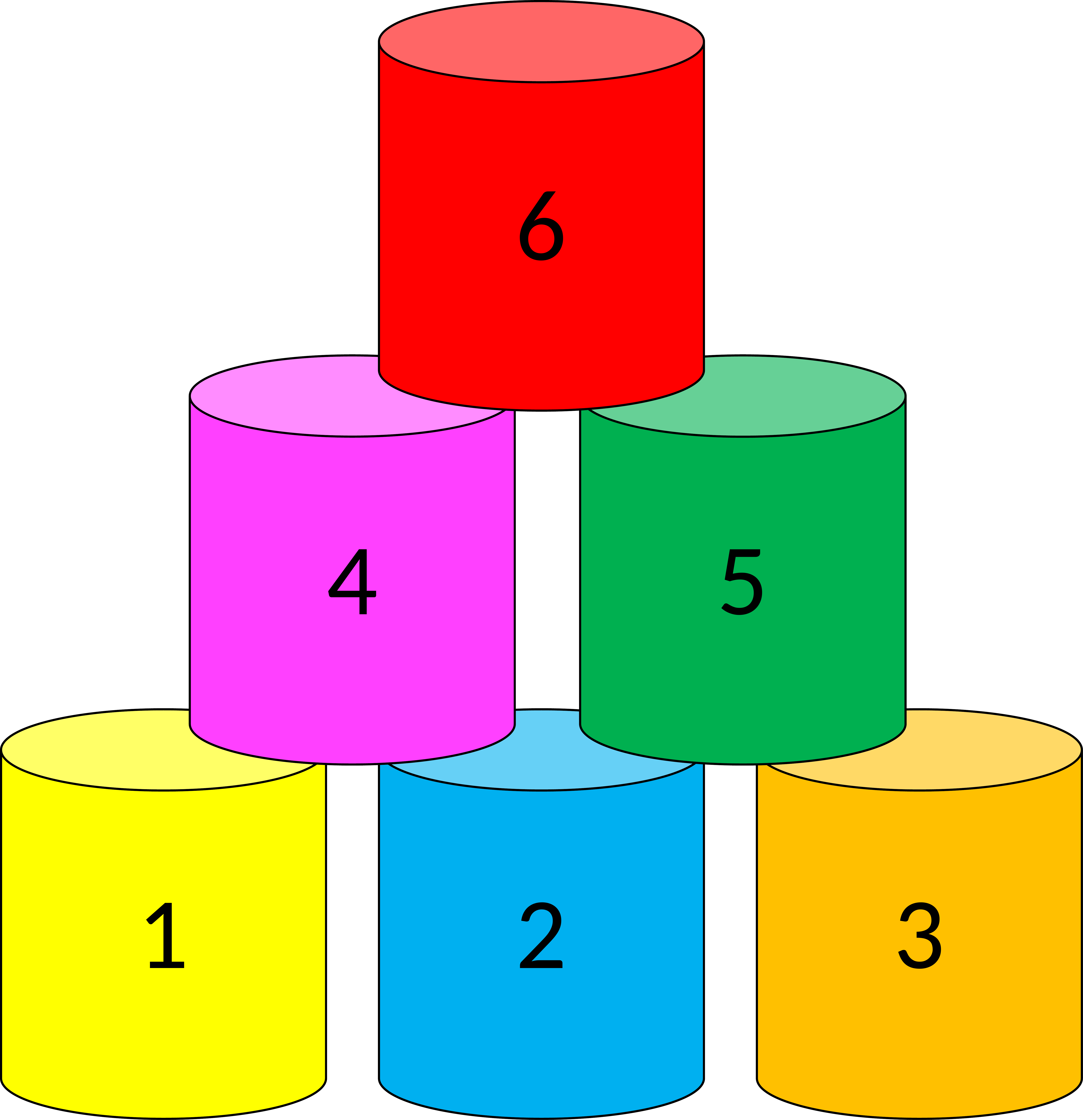}
        \caption{Start pose}
        \label{}
    \end{subfigure}
    \hspace{.1in}
    \begin{subfigure}{.22\textwidth}
    \centering
        \includegraphics[height=1.3in]{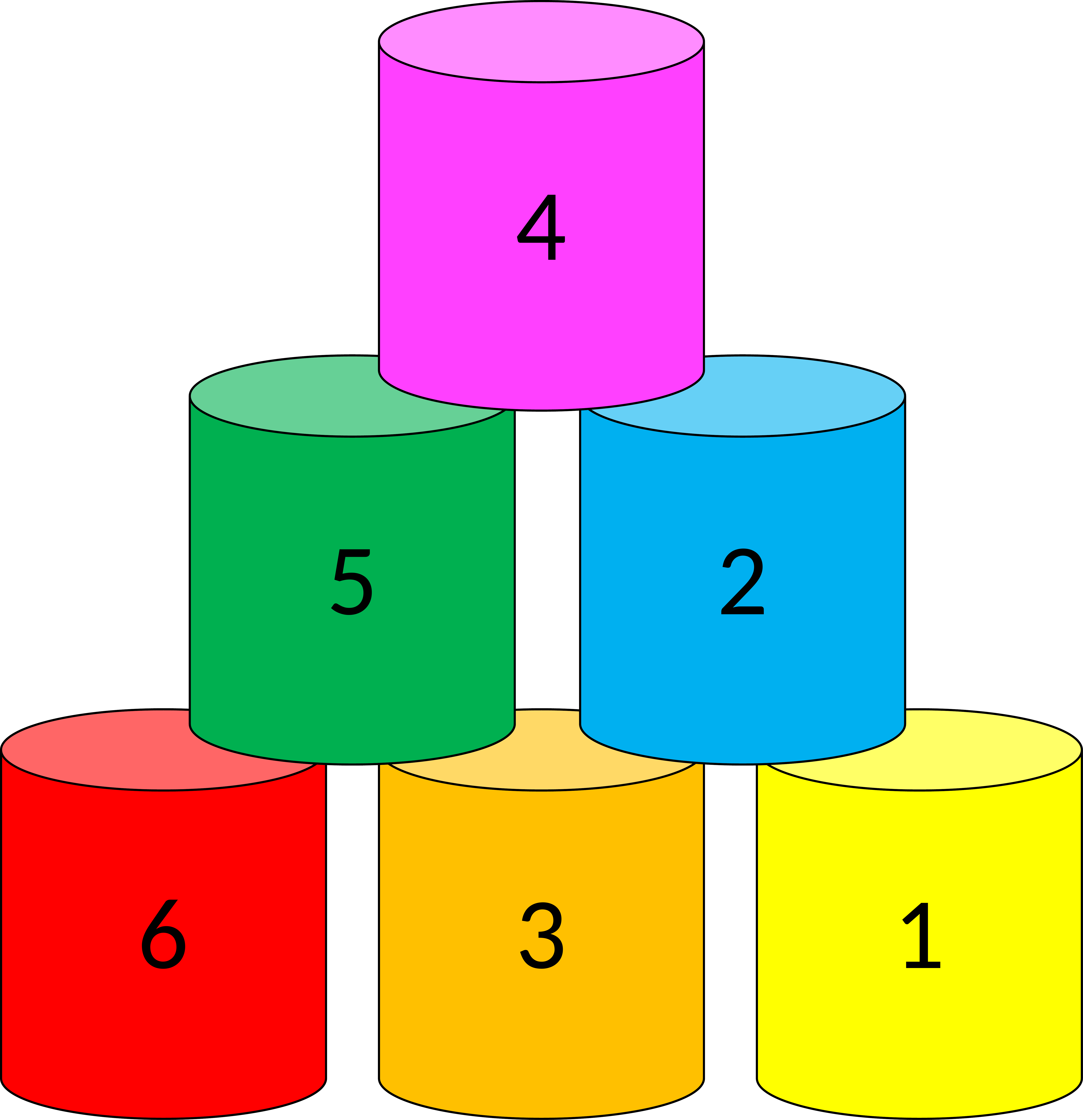}
        \caption{Goal pose}
        \label{}
    \end{subfigure}
    \caption{A ``2D pyramid'' setup with six objects. We use this to show the solution optimality difference between \algo and greedy approaches.  \label{fig:4}}
\end{figure}

As will be shown in Sec.~\ref{sec:evaluation}, computationally, the greedy algorithm is extremely efficient in many cases (but not always). However, it often does not provide the optimal solution. As an illustrative example, for the \probm instance in Fig.~\ref{fig:4}, the greedy algorithm takes $11$ steps as ($b$ means buffer and $g$ means goal for the particular object): $6 \to b, o_4 \to b, o_5 \to b, o_3 \to b, o_1 \to g, o_6 \to g, o_2 \to b, o_3 \to g, o_2 \to g, o_5 \to g, o_4 \to g$. 
On the other hand, \algo gives the optimal $10$ step solution as: $6 \to b, o_5 \to b, o_4 \to b, o_2 \to b, o_3 \to g, o_1 \to g, o_6 \to g, o_2 \to g, o_5 \to g, o_4 \to g$.

We note that the example in Fig.~\ref{fig:4} is a relatively simple case, where the greedy algorithm would do $10\%$ worse than the optimal solution in terms of solution quality.

\section{Evaluation}\label{sec:evaluation}
\begin{figure}[!ht]
\vspace{2mm}
    \centering
        \includegraphics[width=\columnwidth]{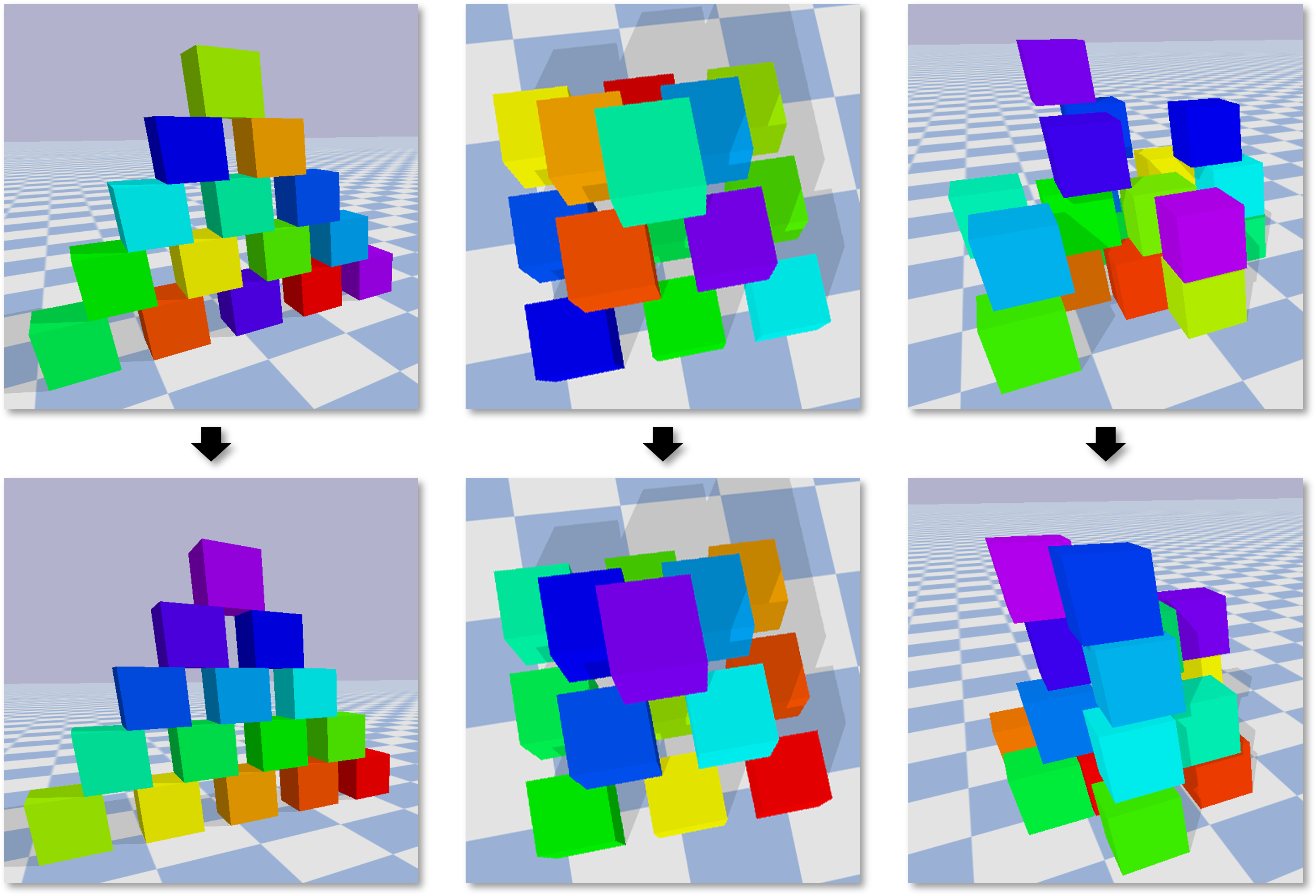}
    \caption{The three types of simulation-based evaluation setups. From left to right: 2D Pyramid, 3D Pyramid, and Random Pile. The top row shows the start configurations and the bottom row shows the corresponding goal configurations.\label{fig:pb}}
\end{figure}

To evaluate the performance of \algo, we conduct numerical experiments under three scenarios (see Fig.~\ref{fig:pb} for examples of each): 
\begin{enumerate}[leftmargin=5mm]
    \item \textbf{2D Pyramid}: In a 2D pyramid with $m$ layers, unit cubes have the same y coordinate. 
    For the $i^{th}$ layer in the pyramid, there are $m-i+1$ cubes. 
    In this scenario, we rearrange objects from one 2D pyramid to another 2D pyramid with the same dimensions. 
    Object labels are randomly selected. 
    Fig.~\ref{fig:pb} [Left] shows an example of 2D Pyramid with $3$ layers.
    
    \item \textbf{3D Pyramid}: Similarly, for a 3D pyramid with $m$ layers, the $i^{th}$ layer has $(m-i+1)^2$ unit cubes.
    In this scenario, we rearrange objects from one 3D pyramid to another 3D pyramid with the same dimensions.
    Again, object labels are random. 
    Fig.~\ref{fig:pb} [Middle] shows an example of 3D Pyramid with $3$ layers.
    \item \textbf{Random Piles}: In this scenario, an object arrangement is generated as follows. 
    We spawn axis-aligned unit cubes with random $x$ and $y$ in a $5\times 5$ square region one after another.
    If the footprint of the new object overlaps with existing objects, it will be generated on top of the overlapping objects.
    The arrangement will be discarded if any generated object pose is unstable.
\end{enumerate}

We note \algo does not require the objects to be cubes (see one of the real robot experiments). In addition to different formations of objects in the workspace, we evaluate algorithm performance on test cases with different positional relationships of start and goal arrangements. An instance is called \emph{disjoint} if the volume occupied by the start arrangement does not intersect the volume occupied by the goal arrangement. Otherwise, the instance is called \emph{in-place}.

Both \algo and the greedy method are implemented in Python. All experiments are executed on an Intel$^\circledR$ i5 CPU at 2.7GHz. 
For solving ILP, Gurobi 9.5.1 is used. 
Each data point is the average of 20 (for random piles) or 30 (for pyramids) test cases minus the unfinished ones, if any, subject to a time limit of 300 seconds per test case.

For 2D/3D Pyramid cases, the regular layered structures ensure that all valid intermediate arrangements are also stable. In other words, scenarios shown in Fig.~\ref{fig:1} and Fig.~\ref{fig:3}, where two objects support each other, cannot happen. Therefore, stability checks are not needed.
%
As such, we compare \algo without stability checks and the greedy baseline. This also allows us to test the scalability of the ILP model in \algo independent of the physics engine. 
For the Random Piles instances where instability during manipulation is possible, we compare the complete \algo with the greedy method.

In evaluating the algorithms, up to three metrics are used. Optimality ratio is the ratio between the number of times objects are relocated buffers by an algorithm compared to the optimal number of movements returned by \algo, which is guaranteed to be optimal. Therefore, this is also $1$ for \algo but never smaller than $1$ for the greedy method. 
The other two metrics are success rates and computational times, which are straightforward to interpret.  
The success rate is exclusively included in randomized instances since these test cases are the only ones that exceed the time limit or are unstable.

\subsection{Simulation Experiments}
\begin{figure}[h]
    \centering
    \includegraphics[width=0.48\textwidth]{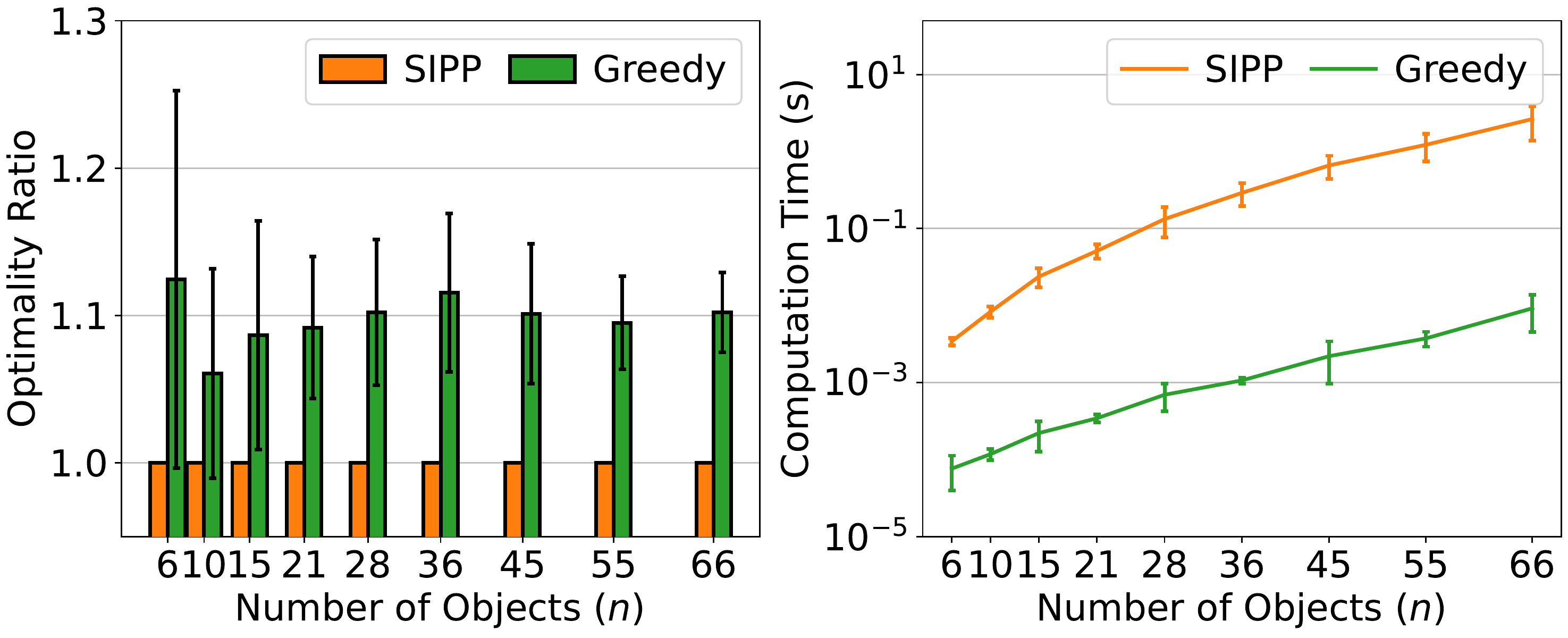}    
    \caption{ Algorithm performances (optimality ratio and computation time) for the in-place 2D Pyramid scenario.}
    \label{fig:in_place_2d}
\end{figure}

\begin{figure}[h]
    \centering
    \includegraphics[width=0.48\textwidth]{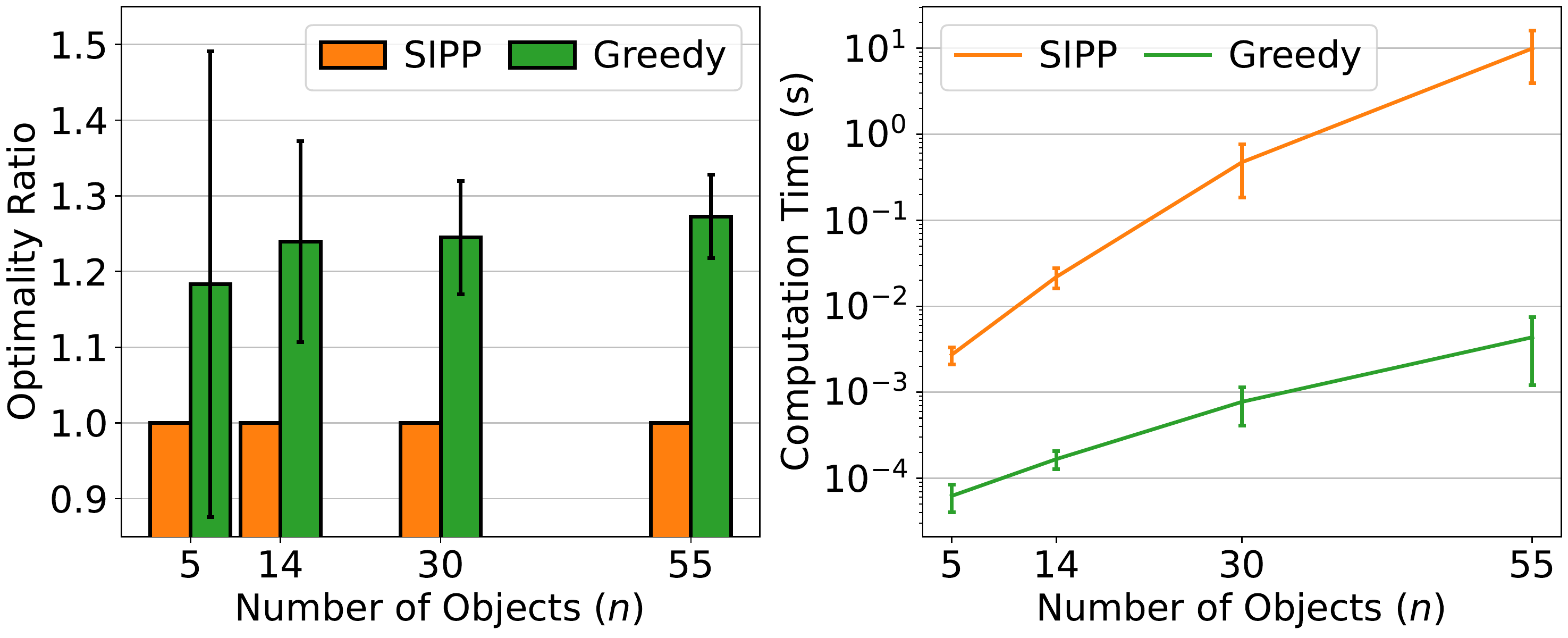}
    \caption{Algorithm performance (optimality ratio and computation time) for the in-place 3D Pyramid scenario.}
    \label{fig:in_place_3d}
\end{figure}

In Fig.~\ref{fig:in_place_2d} and \ref{fig:in_place_3d}, we show the results of \emph{in-place} 2D and 3D Pyramid scenarios.
In 2D Pyramid instances, as shown by the optimality ratio, the saving of \algo slowly increases and reaches around $11\%$ when $n=66$ compared with the greedy method.
In terms of the raw numbers of additional moves, for the $n=66$ case, the greedy method needs $5.5$ extra buffer relocations on average, out of a total of around $60$ relocations to the buffer and around $120$ total moves. That is, \algo provides about $3$-$4\%$ execution time savings. which can be significant depending on the practical application. 
Even in $66$-cube instances, the ILP model in \algo computes optimal rearrangement plans in $4.70$ secs on average.
In in-place 3D Pyramid instances, with the additional combinatorial constraints between layers, \algo saves up to $28\%$ actions moving objects to buffers.
When $n = 55$, the greedy method needs $9$ extra buffer relocations on average, out of a total of $40$ relocations to the buffer and $90$ total moves. This translates to over $10\%$ of execution time savings. This is more than doubling the 2D Pyramid case.
Despite the additional constraints, the ILP model is still scalable, computing optimal rearrangement plans in $5.44$ secs on average for $55$-object instances.
\jy{We need raw data to mention some numbers. Otherwise, we will be questioned on the presentation.}

\begin{figure}[h]
 \vspace{2mm}
    \centering
    
    \includegraphics[width=0.48\textwidth]{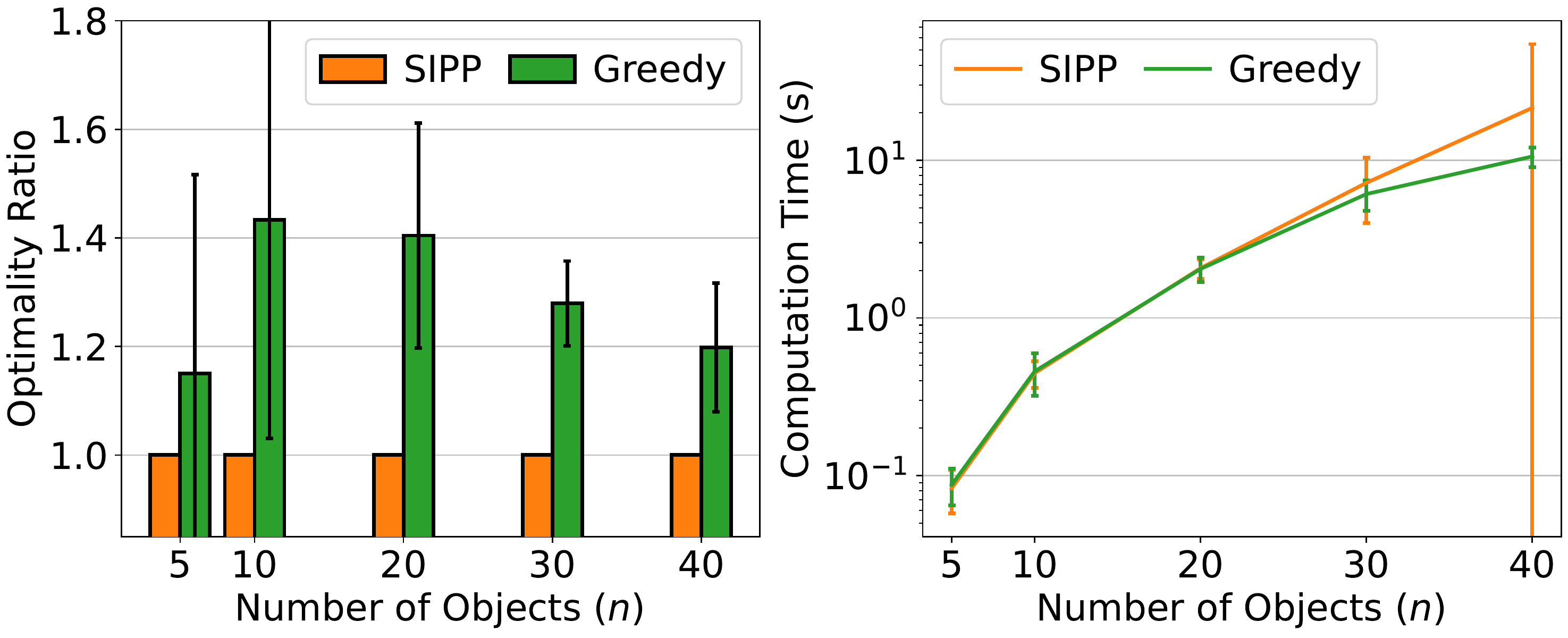}
    \includegraphics[width=0.35\textwidth]{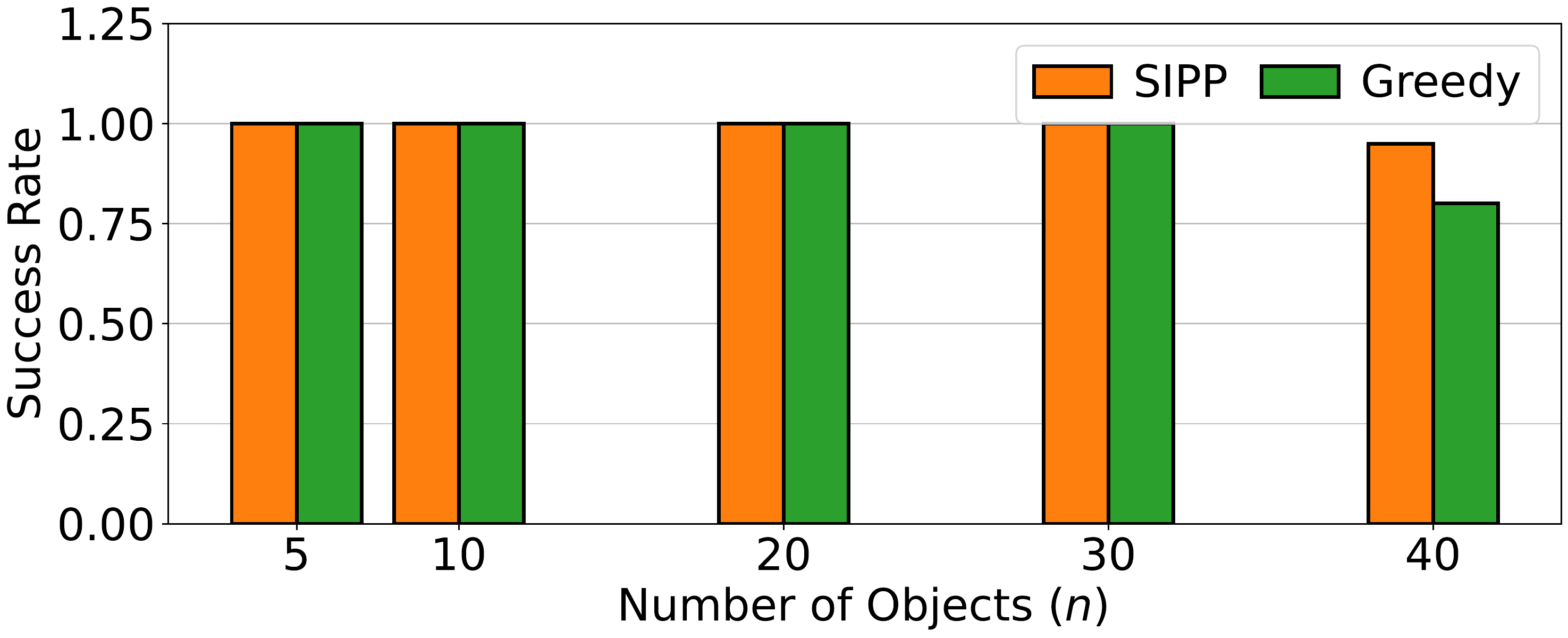}

    \caption{Algorithm performance (optimality ratio, computation time, and success rate) for the in-place Random Pile scenario.}
    \label{fig:in_place_random}
\end{figure}

The Random Pile results, given in Fig.~\ref{fig:in_place_random}, show results similar to the 3D setting. Interestingly, the greedy method does the worst in the mid-range (when $n=10$ and $20$) as the number of objects increases.
For $n=10$ and $20$, each arrangement has only 1-2 layers; in this case, the greedy method has a particularly difficult time getting the right rearrangement order. 
%
When $n=40$, with the stability checker, \algo computes the shortest stable plans in around $30$ secs on average.
In the same cases, the greedy method fails in around $20\%$ cases due to the instability of intermediate arrangements in returned plans.
We note that there isn't a straightforward way to integrate stability checks into the greedy method without significantly degrading its performance due to the incompatibility between stability checks and searching through the combinatorial choices.
On the side of solution optimality, \algo does even better than earlier settings. 

\begin{figure}[h]
    \centering
    \includegraphics[width=0.48\textwidth]{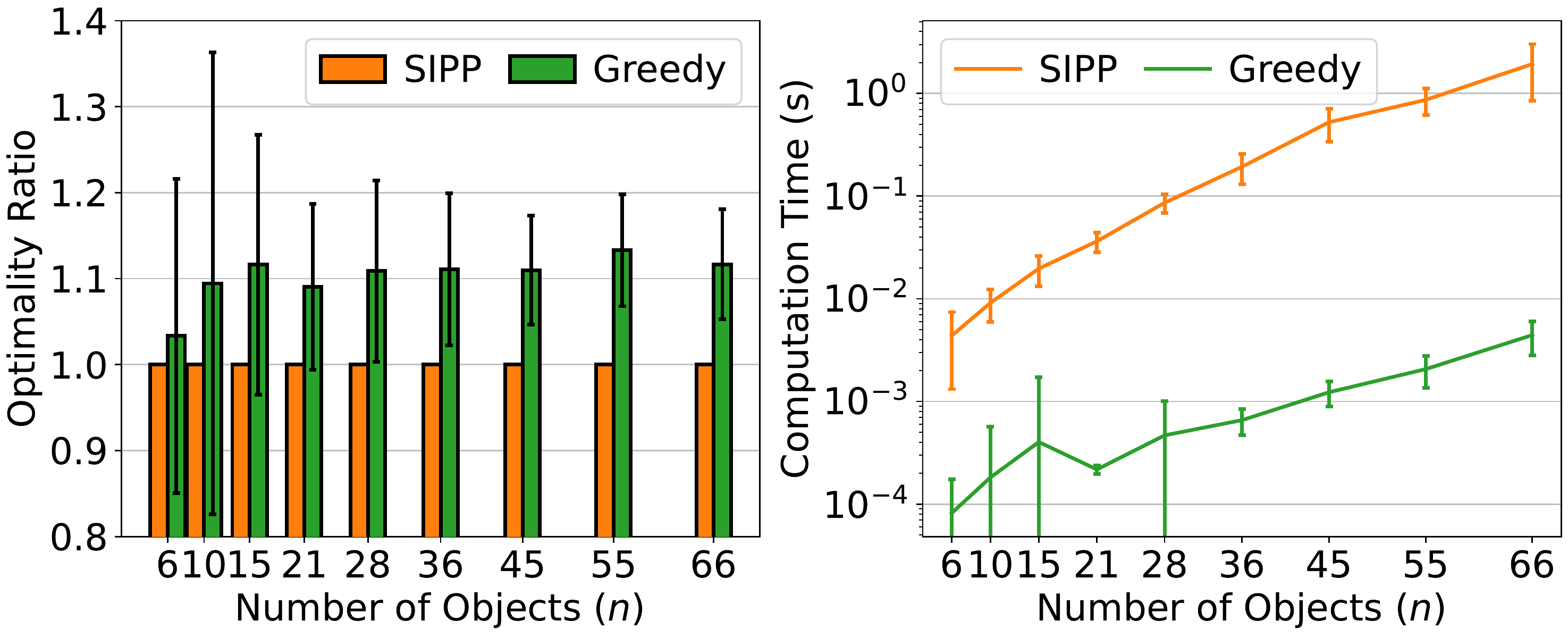}
    \caption{Algorithm performance (optimality ratio and computation time) for the disjoint 2D Pyramid scenario.}
    \label{fig:out_of_place_2d}
\end{figure}

\begin{figure}[h]
\vspace{2mm}
     \centering
    \includegraphics[width=0.48\textwidth]{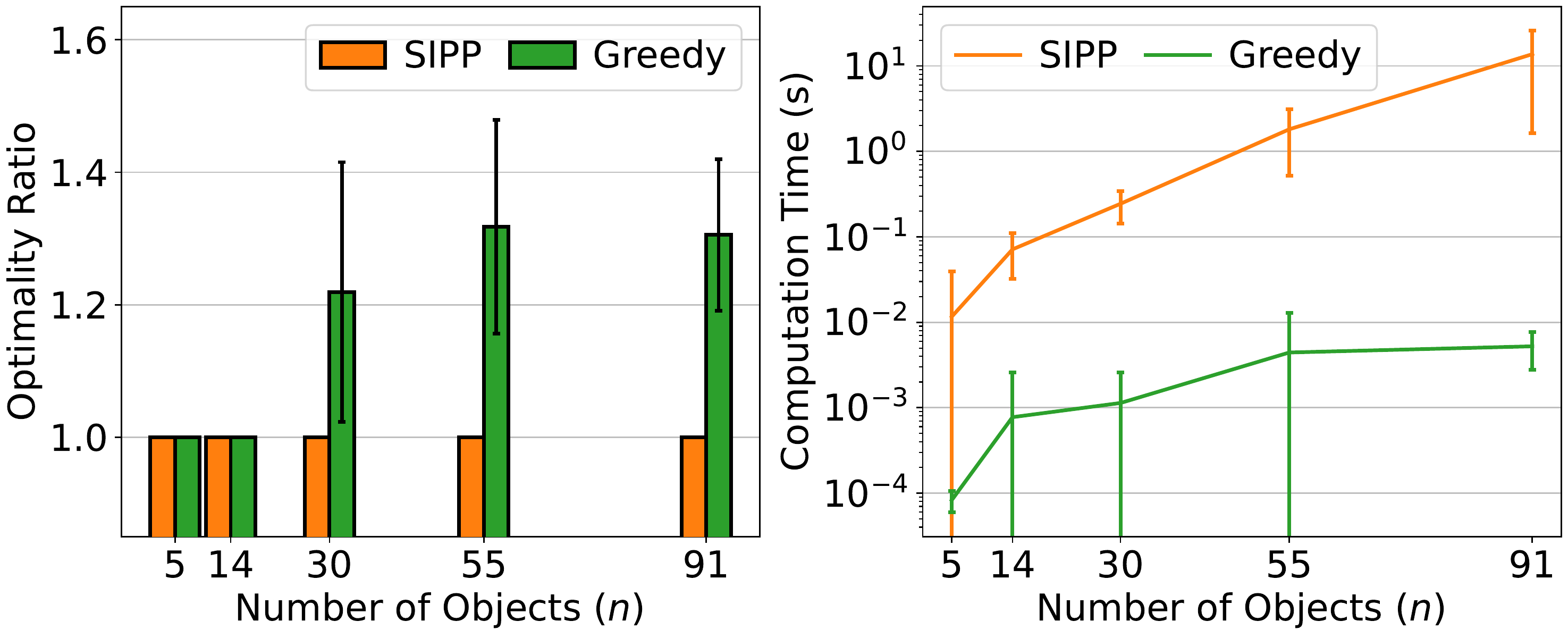}
    \caption{Algorithm performance (optimality ratio and computation time) for disjoint 3D Pyramid scenario.}
    \label{fig:out_of_place_3d}
\end{figure}

\begin{figure}[h]
    \centering
    \includegraphics[width=0.48\textwidth]{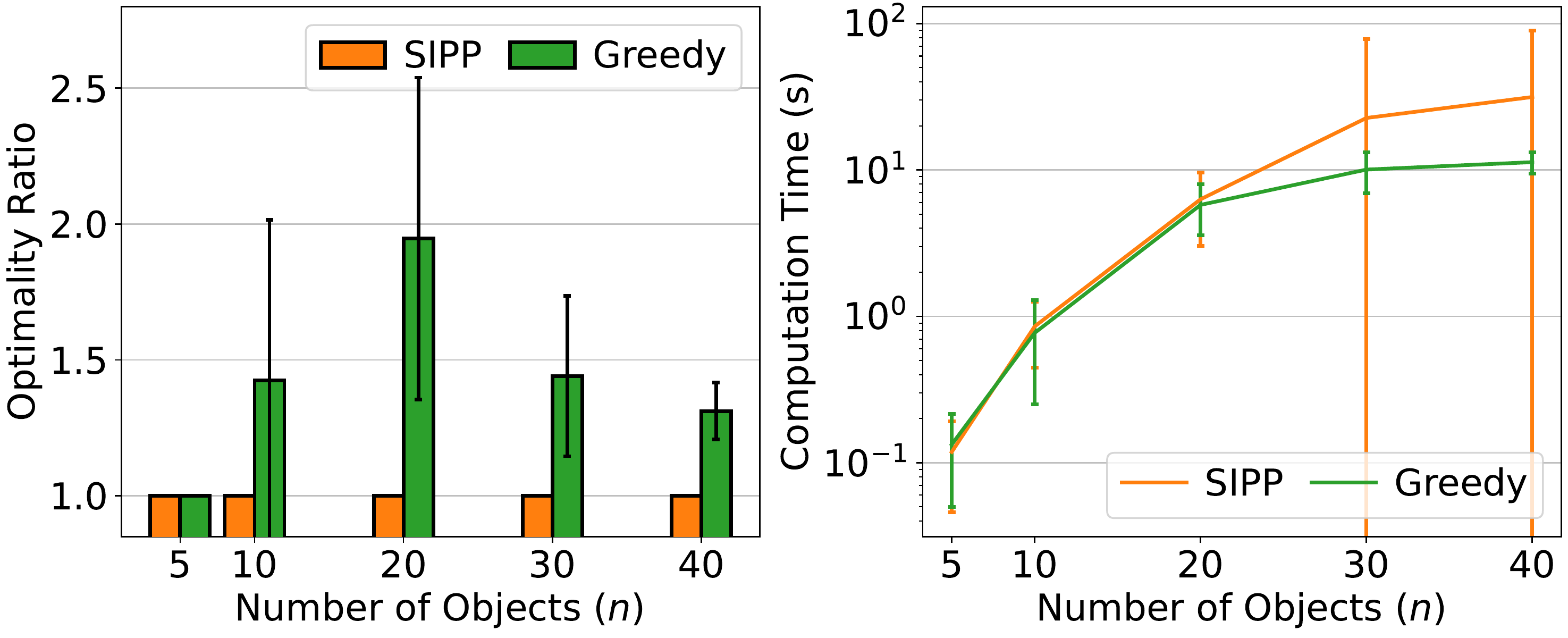}
    \includegraphics[width=0.35\textwidth]{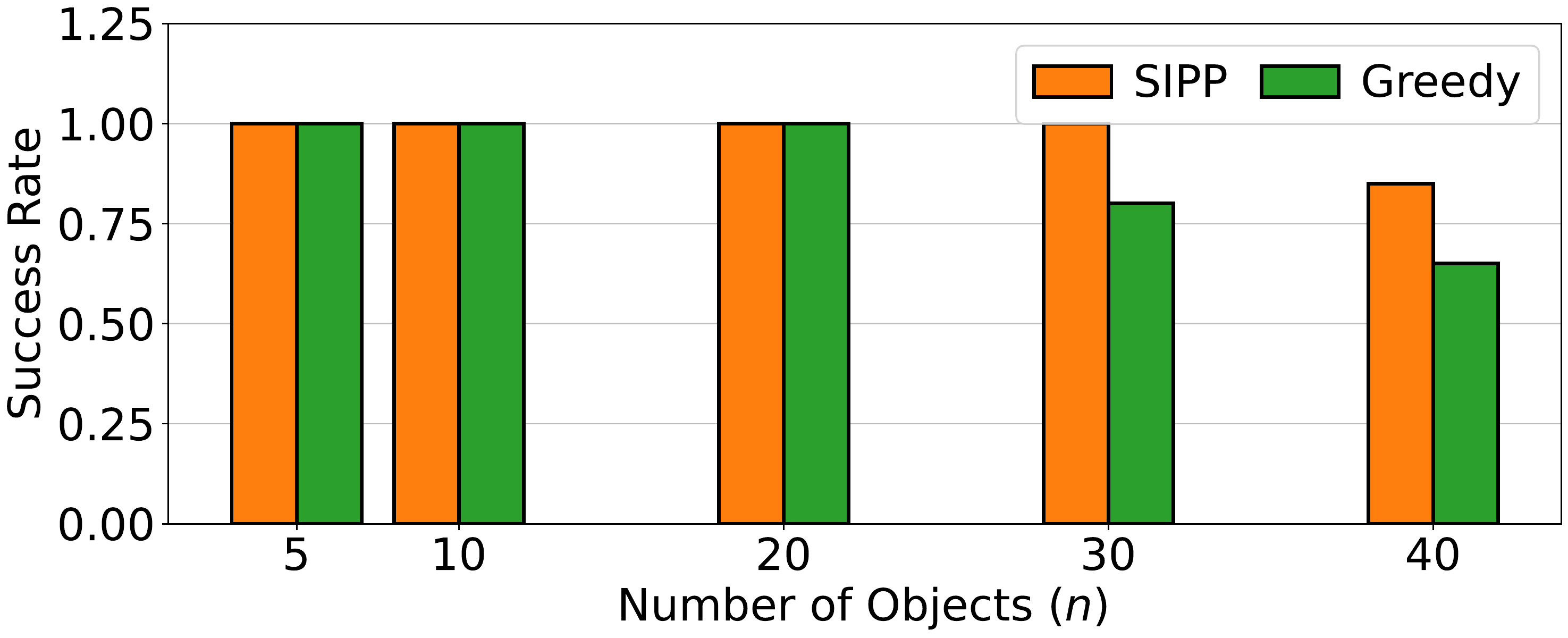}
    \caption{Algorithm performance (optimality ratio, computation time, and success rate) for the disjoint Random Pile scenario.}
    \label{fig:out_of_place_random}
    \vspace{-5mm}
\end{figure}

Fig.~\ref{fig:out_of_place_2d}, \ref{fig:out_of_place_3d}, and \ref{fig:out_of_place_random} show the results for disjoint instances. The optimality ratio is similar in the 2D Pyramid setting but better in the 3D Pyramid and Random Pile settings, though the actual execution time saving again falls between $5$-$10\%$, largely the same as the in-place setting.   
As for computation time, the disjoint 2D Pyramid instances spend significantly lower runtime than their in-place counterparts. 
This makes sense as there are fewer combinatorial constraints for the ILP model to resolve in the disjoint scenario than in the in-place one. 
In fact, Eq.~\eqref{eq:3} will not be used at all in this case since there are no start-goal constraints.
This phenomenon does not occur in 3D pyramid instances, which suggests that the start goal constraints have less impact in this scenario.
In contrast, in Random Pile instances, disjoint instances have more unstable intermediate arrangements so that \algo spends more time on computation and fails to find stable plans in $20\%$ instances when $n=40$.
Furthermore, without the start-goal constraints, the variance in computation time of \algo is much smaller in all three scenarios, especially in problems with a large number of objects, which indicates that the difficulty of the rearrangement problem is greatly affected by the start-goal constraints. 

\subsection{Hardware Demonstrations}
In the accompanying video, we further demonstrate that \algo can be applied to practical, real-world \probm scenarios with objects whose bases are squares and general shapes.
The objects are manipulated with an OnRobot VGC 10 vacuum gripper on a UR-5e robot arm.
Their 3D poses are estimated based on fiducial markers detected by an Intel RealSense D405 RGB-D camera.

\begin{figure}[h!]
    \centering
    \includegraphics[width=0.49\textwidth]{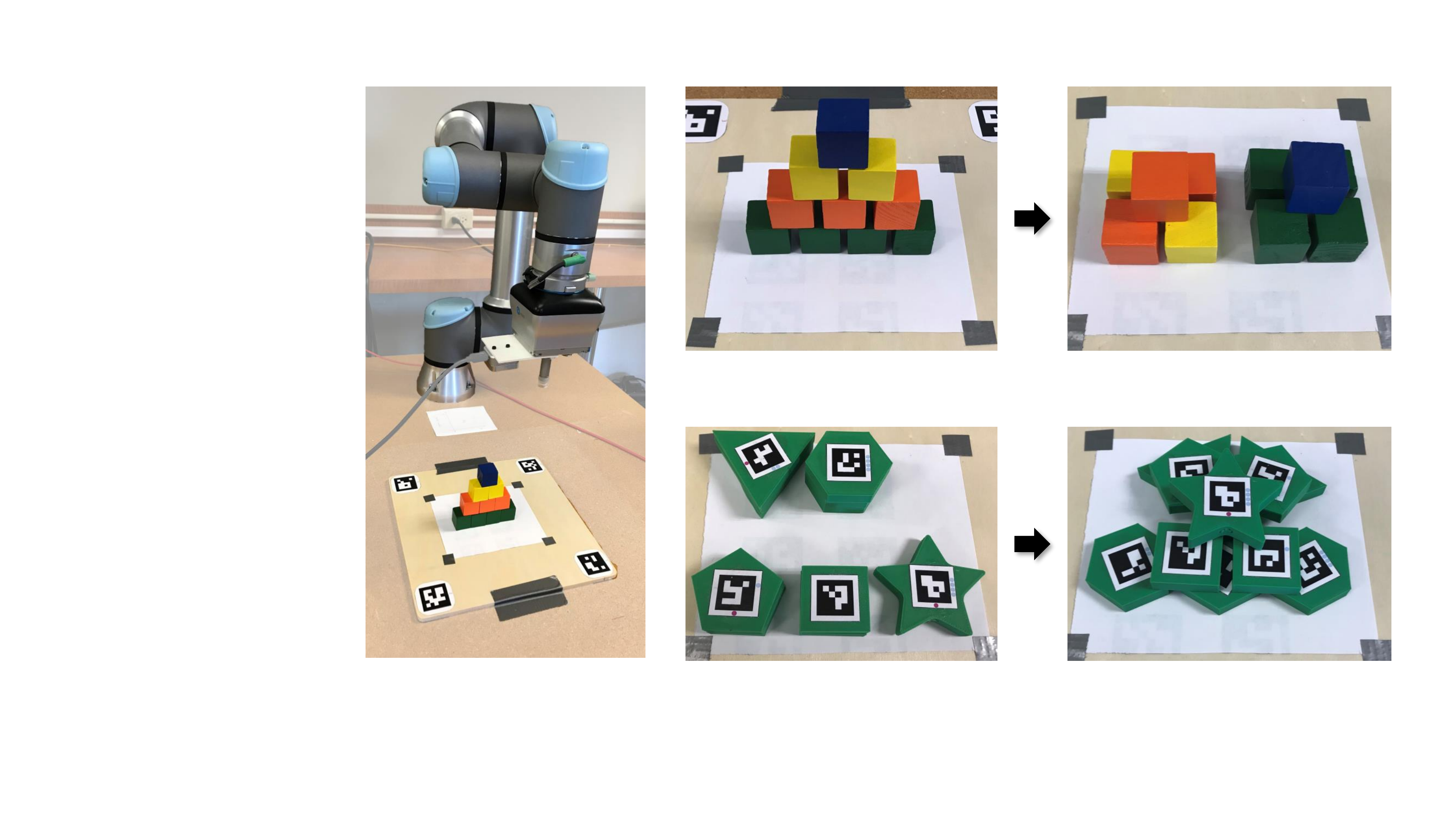}
    \caption{Hardware Demonstrations. \emph{Left}: Our hardware setup for \probm. \emph{Top right}: Rearrangement from a 2D pyramid to two 3D pyramids. \emph{Bottom right}: \probm rearranging objects with general-shaped bases. All objects assume unique IDs.}
    \label{fig:demo}
\end{figure}

\section{Conclusion and Discussions}\label{sec:conclusion}
In this work, we study \probm, a challenging problem of rearranging object piles on a tabletop, where a pile may contain multiple layers of objects.
To our knowledge, our study is the first that examines the multi-layer rearrangement problem from a combinatorial perspective. 
Our proposed solution, \algo, is an optimal planner that interleaves an ILP model and a simulation-based stability checker to handle the combinatorial constraints and structural stability during manipulation.
Extensive experiments demonstrate that \algo outperforms the greedy baseline planner in terms of efficiency and effectively avoids potential instability during the rearrangement.

{\small
\bibliographystyle{formatting/IEEEtran}
\bibliography{bib/bib}

\begin{thebibliography}{10}
\providecommand{\url}[1]{#1}
\csname url@rmstyle\endcsname
\providecommand{\newblock}{\relax}
\providecommand{\bibinfo}[2]{#2}
\providecommand\BIBentrySTDinterwordspacing{\spaceskip=0pt\relax}
\providecommand\BIBentryALTinterwordstretchfactor{4}
\providecommand\BIBentryALTinterwordspacing{\spaceskip=\fontdimen2\font plus
\BIBentryALTinterwordstretchfactor\fontdimen3\font minus
  \fontdimen4\font\relax}
\providecommand\BIBforeignlanguage[2]{{%
\expandafter\ifx\csname l@#1\endcsname\relax
\typeout{** WARNING: IEEEtran.bst: No hyphenation pattern has been}%
\typeout{** loaded for the language `#1'. Using the pattern for}%
\typeout{** the default language instead.}%
\else
\language=\csname l@#1\endcsname
\fi
#2}}

\bibitem{HanStiKonBekYu18IJRR}
\BIBentryALTinterwordspacing
S.~D. Han, N.~M. Stiffler, A.~Krontiris, K.~E. Bekris, and J.~Yu, ``Complexity
  results and fast methods for optimal tabletop rearrangement with overhand
  grasps,'' \emph{The International Journal of Robotics Research}, vol.~37, no.
  13-14, pp. 1775--1795, 2018. [Online]. Available:
  \url{https://doi.org/10.1177/0278364918780999}
\BIBentrySTDinterwordspacing

\bibitem{GaoFenYu21RSS}
K.~Gao, S.~W. Feng, and J.~Yu, ``On minimizing the number of running buffers
  for tabletop rearrangement,'' in \emph{Robotics: Science and Systems
  {(RSS)}}, 2021.

\bibitem{GaoLauHuaBekYu22ICRA}
K.~Gao, D.~Lau, B.~Huang, K.~E. Bekris, and J.~Yu, ``Fast high-quality tabletop
  rearrangement in bounded workspace,'' in \emph{2022 International Conference
  on Robotics and Automation (ICRA)}, 2022, pp. 1961--1967.

\bibitem{gao2023minimizing}
K.~Gao, S.~W. Feng, B.~Huang, and J.~Yu, ``Minimizing running buffers for
  tabletop object rearrangement: Complexity, fast algorithms, and
  applications,'' \emph{The International Journal of Robotics Research}, p.
  02783649231178565, 2023.

\bibitem{9868234}
V.~N. Hartmann, A.~Orthey, D.~Driess, O.~S. Oguz, and M.~Toussaint,
  ``Long-horizon multi-robot rearrangement planning for construction
  assembly,'' \emph{IEEE Transactions on Robotics}, vol.~39, no.~1, pp.
  239--252, 2023.

\bibitem{GaoYu22IROS}
K.~Gao and Jingjin, ``Toward efficient task planning for dual-arm tabletop
  object rearrangement,'' in \emph{2022 IEEE/RSJ International Conference on
  Intelligent Robots and Systems (IROS)}, 2022.

\bibitem{danielczuk2021object}
M.~Danielczuk, A.~Mousavian, C.~Eppner, and D.~Fox, ``Object rearrangement
  using learned implicit collision functions,'' in \emph{2021 IEEE
  International Conference on Robotics and Automation (ICRA)}.\hskip 1em plus
  0.5em minus 0.4em\relax IEEE, 2021, pp. 6010--6017.

\bibitem{ShoSolYuHalBek18WAFR}
R.~Shome, K.~Solovey, J.~Yu, D.~Halperin, and K.~Bekris, ``Fast, high-quality
  dual-arm rearrangement in synchronous, monotone tabletop setups,'' in
  \emph{Proceedings Workshop on Algorithmic Foundations of Robotics {(WAFR)}},
  2018.

\bibitem{WanGaoYuBek22ICAPS}
\BIBentryALTinterwordspacing
R.~Wang, K.~Gao, J.~Yu, and K.~Bekris, ``Lazy rearrangement planning in
  confined spaces,'' vol.~32, no.~1, Jun. 2022, pp. 385--393. [Online].
  Available: \url{https://ojs.aaai.org/index.php/ICAPS/article/view/19824}
\BIBentrySTDinterwordspacing

\bibitem{wang2022efficient}
R.~Wang, Y.~Miao, and K.~E. Bekris, ``Efficient and high-quality prehensile
  rearrangement in cluttered and confined spaces,'' in \emph{2022 International
  Conference on Robotics and Automation (ICRA)}.\hskip 1em plus 0.5em minus
  0.4em\relax IEEE, 2022, pp. 1968--1975.

\bibitem{wada2022reorientbot}
K.~Wada, S.~James, and A.~J. Davison, ``Reorientbot: Learning object
  reorientation for specific-posed placement,'' in \emph{2022 International
  Conference on Robotics and Automation (ICRA)}.\hskip 1em plus 0.5em minus
  0.4em\relax IEEE, 2022, pp. 8252--8258.

\bibitem{gu2022multi}
J.~Gu, D.~S. Chaplot, H.~Su, and J.~Malik, ``Multi-skill mobile manipulation
  for object rearrangement,'' in \emph{Deep Reinforcement Learning Workshop
  NeurIPS 2022}.

\bibitem{gan2022threedworld}
C.~Gan, S.~Zhou, J.~Schwartz, S.~Alter, A.~Bhandwaldar, D.~Gutfreund, D.~L.
  Yamins, J.~J. DiCarlo, J.~McDermott, A.~Torralba, \emph{et~al.}, ``The
  threedworld transport challenge: A visually guided task-and-motion planning
  benchmark towards physically realistic embodied ai,'' in \emph{2022
  International Conference on Robotics and Automation (ICRA)}.\hskip 1em plus
  0.5em minus 0.4em\relax IEEE, 2022, pp. 8847--8854.

\bibitem{szot2021habitat}
A.~Szot, A.~Clegg, E.~Undersander, E.~Wijmans, Y.~Zhao, J.~Turner, N.~Maestre,
  M.~Mukadam, D.~S. Chaplot, O.~Maksymets, \emph{et~al.}, ``Habitat 2.0:
  Training home assistants to rearrange their habitat,'' \emph{Advances in
  Neural Information Processing Systems}, vol.~34, pp. 251--266, 2021.

\bibitem{huang2019large}
E.~Huang, Z.~Jia, and M.~T. Mason, ``Large-scale multi-object rearrangement,''
  in \emph{2019 International Conference on Robotics and Automation
  (ICRA)}.\hskip 1em plus 0.5em minus 0.4em\relax IEEE, 2019, pp. 211--218.

\bibitem{Han21CASE}
S.~D. Han, B.~Huang, S.~Ding, C.~Song, S.~W. Feng, M.~Xu, H.~Lin, Q.~Zou,
  A.~Boularias, and J.~Yu, ``Toward fully automated metal recycling using
  computer vision and non-prehensile manipulation,'' in \emph{2021 IEEE 17th
  International Conference on Automation Science and Engineering (CASE)}, 2021,
  pp. 891--898.

\bibitem{song2020multi}
H.~Song, J.~A. Haustein, W.~Yuan, K.~Hang, M.~Y. Wang, D.~Kragic, and J.~A.
  Stork, ``Multi-object rearrangement with monte carlo tree search: A case
  study on planar nonprehensile sorting,'' in \emph{2020 IEEE/RSJ International
  Conference on Intelligent Robots and Systems (IROS)}.\hskip 1em plus 0.5em
  minus 0.4em\relax IEEE, 2020, pp. 9433--9440.

\bibitem{HuaHanBouYu21ICRA}
B.~Huang, S.~D. Han, A.~Boularias, and J.~Yu, ``{DIPN: Deep Interaction
  Prediction Network with Application to Clutter Removal},'' in \emph{2021 IEEE
  International Conference on Robotics and Automation (ICRA)}, 2021, pp.
  4694--4701.

\bibitem{gao2022utility}
K.~Gao and J.~Yu, ``On the utility of buffers in pick-n-swap based lattice
  rearrangement,'' \emph{arXiv preprint arXiv:2209.05390}, 2022.

\bibitem{HanStiBekYu18RAL}
S.~D. Han, N.~M. Stiffler, K.~E. Bekris, and J.~Yu, ``Efficient, high-quality
  stack rearrangement,'' \emph{IEEE Robotics and Automation Letters}, vol.~3,
  no.~3, pp. 1608--1615, 2018, note: presented at ICRA 2018.

\bibitem{SzeYu21SPAR}
M.~Szegedy and J.~Yu, ``On rearrangement of items stored in stacks,'' in
  \emph{Algorithmic Foundations of Robotics XIV}, S.~M. LaValle, M.~Lin,
  T.~Ojala, D.~Shell, and J.~Yu, Eds.\hskip 1em plus 0.5em minus 0.4em\relax
  Cham: Springer International Publishing, 2021, pp. 518--533.

\bibitem{russell2010artificial}
S.~J. Russell, \emph{Artificial intelligence a modern approach}.\hskip 1em plus
  0.5em minus 0.4em\relax Pearson Education, Inc., 2010.

\bibitem{fikes1971strips}
R.~E. Fikes and N.~J. Nilsson, ``Strips: A new approach to the application of
  theorem proving to problem solving,'' \emph{Artificial intelligence}, vol.~2,
  no. 3-4, pp. 189--208, 1971.

\bibitem{aeronautiques1998pddl}
C.~Aeronautiques, A.~Howe, C.~Knoblock, I.~D. McDermott, A.~Ram, M.~Veloso,
  D.~Weld, D.~W. SRI, A.~Barrett, D.~Christianson, \emph{et~al.}, ``Pddl| the
  planning domain definition language,'' \emph{Technical Report, Tech. Rep.},
  1998.

\bibitem{garrett2020pddlstream}
C.~R. Garrett, T.~Lozano-P{\'e}rez, and L.~P. Kaelbling, ``Pddlstream:
  Integrating symbolic planners and blackbox samplers via optimistic adaptive
  planning,'' in \emph{Proceedings of the International Conference on Automated
  Planning and Scheduling}, vol.~30, 2020, pp. 440--448.

\bibitem{de1987simplified}
T.~De~Fazio and D.~Whitney, ``Simplified generation of all mechanical assembly
  sequences,'' \emph{IEEE Journal on Robotics and Automation}, vol.~3, no.~6,
  pp. 640--658, 1987.

\bibitem{de1989correct}
L.~H. De~Mello and A.~C. Sanderson, ``A correct and complete algorithm for the
  generation of mechanical assembly sequences,'' in \emph{1989 IEEE
  International Conference on Robotics and Automation}.\hskip 1em plus 0.5em
  minus 0.4em\relax IEEE Computer Society, 1989, pp. 56--57.

\bibitem{de1990and}
------, ``And/or graph representation of assembly plans,'' \emph{IEEE
  Transactions on robotics and automation}, vol.~6, no.~2, pp. 188--199, 1990.

\bibitem{wilson1994geometric}
R.~H. Wilson and J.-C. Latombe, ``Geometric reasoning about mechanical
  assembly,'' \emph{Artificial Intelligence}, vol.~71, no.~2, pp. 371--396,
  1994.

\bibitem{jimenez2013survey}
P.~Jim{\'e}nez, ``Survey on assembly sequencing: a combinatorial and
  geometrical perspective,'' \emph{Journal of Intelligent Manufacturing},
  vol.~24, pp. 235--250, 2013.

\bibitem{wan2018assembly}
W.~Wan, K.~Harada, and K.~Nagata, ``Assembly sequence planning for motion
  planning,'' \emph{Assembly Automation}, vol.~38, no.~2, pp. 195--206, 2018.

\bibitem{dobashi2014robust}
H.~Dobashi, J.~Hiraoka, T.~Fukao, Y.~Yokokohji, A.~Noda, H.~Nagano,
  T.~Nagatani, H.~Okuda, and K.-i. Tanaka, ``Robust grasping strategy for
  assembling parts in various shapes,'' \emph{Advanced Robotics}, vol.~28,
  no.~15, pp. 1005--1019, 2014.

\bibitem{mcevoy2014assembly}
M.~McEvoy, E.~Komendera, and N.~Correll, ``Assembly path planning for stable
  robotic construction,'' in \emph{2014 IEEE International Conference on
  Technologies for Practical Robot Applications (TePRA)}.\hskip 1em plus 0.5em
  minus 0.4em\relax IEEE, 2014, pp. 1--6.

\bibitem{dogar2019multi}
M.~Dogar, A.~Spielberg, S.~Baker, and D.~Rus, ``Multi-robot grasp planning for
  sequential assembly operations,'' \emph{Autonomous Robots}, vol.~43, pp.
  649--664, 2019.

\bibitem{noseworthy2021active}
M.~Noseworthy, C.~Moses, I.~Brand, S.~Castro, L.~Kaelbling,
  T.~Lozano-P{\'e}rez, and N.~Roy, ``Active learning of abstract plan
  feasibility,'' \emph{arXiv preprint arXiv:2107.00683}, 2021.

\bibitem{garrett2020scalable}
C.~R. Garrett, Y.~Huang, T.~Lozano-P{\'e}rez, and C.~T. Mueller, ``Scalable and
  probabilistically complete planning for robotic spatial extrusion,''
  \emph{arXiv preprint arXiv:2002.02360}, 2020.

\bibitem{coumans2019}
E.~Coumans and Y.~Bai, ``Pybullet, a python module for physics simulation for
  games, robotics and machine learning,'' \url{http://pybullet.org},
  2016--2019.

\end{thebibliography}
}

\end{document}